%% file: main.tex
\pdfoutput=1

\documentclass[11pt]{article}
\usepackage[]{acl}
\usepackage{times}
\usepackage{latexsym}
\usepackage[T1]{fontenc}
\usepackage[utf8]{inputenc}
\usepackage{microtype}
\usepackage{inconsolata}
\usepackage{graphicx}
\usepackage{multirow}
\usepackage{xspace}
\usepackage{booktabs}
\usepackage{amsmath}
\usepackage{subcaption}
\usepackage{enumitem}
\usepackage{bm}
\definecolor{customgray}{HTML}{eeefee} % 定义颜色
\usepackage{url}
\usepackage{amssymb}
\usepackage{tabularx,ulem}
\usepackage[dvipsnames]{xcolor}
\usepackage[most]{tcolorbox}
\definecolor{promptbox_color_1}{HTML}{1F4E79} % deep navy   — construction
\definecolor{promptbox_color_2}{HTML}{2C7A7B} % teal         — evaluation
\definecolor{promptbox_color_3}{HTML}{9B2C2C} % crimson      — judge

\usepackage{listings}
  \lstdefinestyle{nmqaprompt}{basicstyle=\ttfamily\scriptsize,
      breaklines=true, breakatwhitespace=true, columns=fullflexible,
      frame=single, framesep=4pt, xleftmargin=4pt, xrightmargin=4pt,
      aboveskip=4pt, belowskip=6pt, keepspaces=true, showstringspaces=false}

\usepackage{pifont}
\usepackage{colortbl}
\usepackage{array}

\newcommand{\benchname}{NextMotionQA}
\definecolor{c_id}{HTML}{2D6CC0}   % Identify (locate) 
\definecolor{c_rec}{HTML}{6E9B3E}  % Token Recall      
\definecolor{c_cor}{HTML}{C0392B}  % Correct (fix)      
\definecolor{goldc}{RGB}{12,124,72}   % correct answer / target attribute
\definecolor{fixc}{RGB}{20,90,200}    % model-target correction
\definecolor{errc}{RGB}{200,40,40}    % corrupted (wrong) token
\newcommand{\gold}[1]{\textbf{\textcolor{goldc}{#1}}}
\newcommand{\fix}[1]{\textbf{\textcolor{fixc}{#1}}}
\newcommand{\err}[1]{\textcolor{errc}{\sout{#1}}}
\newcommand{\cmark}{\textcolor{ForestGreen!80!black}{\ding{51}}}
\newcommand{\xmark}{\textcolor{red!60!black}{\ding{55}}}
\newcommand{\pmark}{\textcolor{orange!90!black}{$\bullet$}} % partial

\title{\benchname{}: Benchmarking and Judging Human Motion Understanding with Vision-Language Models}

\newcommand{\tubingen}{$^1$}
\newcommand{\saar}{$^2$}
\newcommand{\tubingensaar}{$^{1,2}$}
\author{Yong Cao\tubingen, Chuqiao Li\tubingen, Xianghui Xie\tubingensaar, Gerard Pons-Moll\tubingensaar, Andreas Geiger\tubingen
\\
{\tubingen}University of Tübingen, Tübingen AI Center, Germany \\
{\saar}Max Planck Institute for Informatics, Saarland Informatics Campus, Germany \\
\texttt{ yong.cao@uni-tuebing.de}
}

\begin{document}
\maketitle

\begin{abstract}
    Reliable evaluation of human motion understanding is fundamental to advancing embodied AI, robotics, and animation. However, existing benchmarks suffer from coarse semantic granularity, undifferentiated difficulty, limited annotation quality, and pervasive answer ambiguity, leaving them unable to diagnose where current models fail. To bridge this gap, we introduce \benchname{}, a comprehensive benchmark that leverages vision-language models (VLMs) for semi-automated, expert-verified dataset. \benchname{} features three complementary tasks: multiple-choice question answering, video captioning, and fine-grained error correction. Each task is systematically structured across three core semantic axes and stratified into three task complexity levels. Our extensive evaluation of twelve representative VLMs uncovers critical capability gaps and weakness that remain invisible under conventional, single-task evaluations. In a complementary direction, recent work has begun using VLMs as judges for text-to-motion evaluation; we ask whether they show the same degradation under harder tasks. We find that VLMs align strongly with expert ratings on coarse criteria (Cohen's $\kappa=0.70$) but break down on fine-grained, part-level judgment ($\kappa=0.10$), validating the paradigm in its strong regime while clarifying its limits. We will release our code and data upon publication.
\end{abstract}

\input{section/introduction}
\input{section/rw}
\input{section/pilot_study}

\input{section/benchmark}

\input{section/experiment}

\input{section/conclusion}
\bibliography{anthology, custom}
\input{appendix}

\end{document}

%% file: section/introduction.tex
\begin{figure*}[t]
    \centering
    \includegraphics[width=\linewidth]{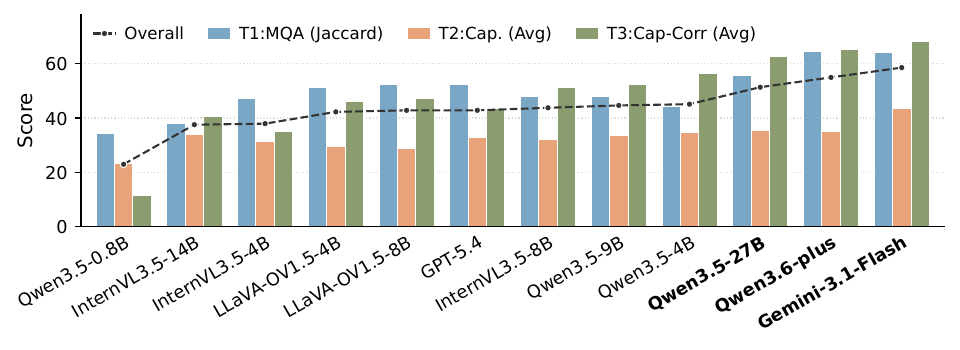}
    \caption{%
        Overall ranking of evaluated VLMs on our \benchname{}, sorted by the mean of Task 1: Multiple-choice Question Answering (MQA), Task 2: Caption (Cap.), and Task 3: Caption Error Correction (Cap-Corr).
        Frontier models (Gemini-3.1-Flash, Qwen3.6-plus) lead overall, while
        Qwen3.5-27B is the strongest fully open-source model. Note the
        anomalous behavior of InternVL3.5-14B, which underperforms its
        smaller variants on MQA. 
        % % \XH{T1,T2,T3 is very confusing here. You can use accron in the figure, and explain in caption what each eval QA set is. e.g. Overall ranking of evaluated VLMs on our \benchname{}, sorted by the mean of three distinct evaluation sets: multiple choice questions (MCQ), free-form captioning(FF-Cap), and caption error correction (Cap-Corr) (or other acron names). 
        % }
    }
    \label{fig:overall_ranking}
\end{figure*}

\section{Introduction}

Human motion plays a central role across a range of vision-and-language research directions, including text-to-motion generation (T2M) \citep{tevet2023mdm, jiang2023motiongpt, guo2024momask}, vision-language-action models for embodied agents \citep{brohan2023rt2, kim2024openvla, black2024pi0}, and motion-aware multimodal dialogue \citep{chen2024motionllm, zhou2024avatargpt}. Reliably evaluating how well models understand human motion is therefore a prerequisite for progress in all of these directions toward training stronger models.

Despite recent attempts \cite{hong2025motionbench, tu2025favor} to construct benchmarks for human motion understanding, existing efforts exhibit three interrelated weaknesses: coarse semantic granularity, the absence of difficulty stratification, and limited annotation quality. Taking HumanMotionQA \citep{endo2023motion} for example, our pilot study (\S\ref{sec:pilot}) shows that three domain experts agree only on $34.3\%$ of the sampled gold answers and mean human accuracy ($52.9\%$) is even worse than public motion-specific model~\citep{endo2023motion} at $57.8\%$. 
% Taking HumanMotionQA \citep{endo2023motion} as a starting point, we conduct a pilot study (\S\ref{sec:pilot}) that makes the third issue concrete: three motion-domain experts agree on the gold answer in only $34.3\%$ of sampled items and reach $52.9\%$ mean accuracy, which is already exceeded by a published motion-specific model at $57.8\%$. 
When domain experts cannot agree on the labels that models produce, reported accuracy reflects label noise rather than genuine model capability.
Recent benchmarks \citep{chen2024motionllm} broaden the input modality to real-world video but retain a QA-only task without explicit difficulty stratification, leaving the same diagnostic deficit.

To address these limitations, we introduce \benchname{}, a $3{\times}3{\times}3$ benchmark of 1{,}307 expert-verified instances that jointly varies three task formats (multiple-choice QA, free-form captioning, and fine-grained error correction), three semantic axes (body-part, direction, action), and three difficulty tiers, with instances drafted by frontier VLMs and filtered under a rejection rubric via human experts derived from our pilot study. Benchmarking twelve representative systems
spanning open-source and frontier close source VLMs, we obtain a clean ranking that separates model families and scales (see Figure~\ref{fig:overall_ranking}), and decomposing the result along the matrix exposes three patterns flat metrics would hide: no system dominates across all three formats, translation direction is the universal weak axis, and performance degrades obviously as difficulty rises.

The sharp degradation we observe on harder motion-understanding tasks raises a natural follow-up: does the same pattern carry over when VLMs are used as evaluators rather than as systems under test? This question is increasingly consequential, as LLM- and VLM-as-a-judge protocols have become common for open-ended evaluation \citep{zheng2023judging, liu2023geval, chen2024mllmjudge} and have recently been adopted for text-to-motion generation, where standard feature-space metrics \citep{Guo_2022_CVPR} correlate weakly with human perception. We address this by re-purposing the evaluated VLMs as judges of generated motions and validating their outputs against expert ratings. Experimenting on nine T2M methods and 60 human evaluators, agreement is strong on coarse criteria (Cohen's $\kappa = 0.70$) but collapses on fine-grained, part-level judgment ($\kappa = 0.10$). This pattern validates the paradigm in its strong regime while delineating where it currently fails.

In summary, our contributions are threefold:

\begin{itemize}
    \item We conduct a \textbf{pilot study} with motion-domain experts that surfaces the structural failure modes limiting existing human-motion evaluation benchmarks.
    \item We propose \textbf{\benchname{}}, a $3{\times}3{\times}3$ benchmark of $1{,}307$ expert-verified cases, on which we evaluate twelve representative VLMs and characterize their capability gaps along tasks, semantic axis, and difficulty tiers.
    \item We present the first difficulty-stratified \textbf{analysis of VLM-as-a-judge} for T2M generation, validating the paradigm at coarse granularity while documenting where it breaks down.
\end{itemize}

%% file: section/rw.tex
\section{Related Work}

\paragraph{Text-to-motion generation.}
Recent text-to-motion (T2M) research has rapidly diversified along several distinct lines. Diffusion-based systems such as MDM~\citep{tevet2023mdm} and latent motion diffusion (MLD)~\citep{chen2023mld} establish continuous motion synthesis, with later variants like ReMoDiffuse~\citep{zhang2023remodiffuse} and MotionLCM~\citep{dai2024motionlcm} improving diversity, controllability, and sampling efficiency. Discrete-token approaches~\citep{zhang2023generating, jiang2023motiongpt, guo2024momask} cast motion as a language-like sequence, with MotionGPT~\citep{jiang2023motiongpt} further coupling motion tokens to a language model backbone. A more recent line moves to continuous latent spaces to avoid quantization loss, including MotionStreamer~\citep{xiao2025motionstreamer}, MARDM~\citep{meng2025rethinking}, and ActionPlan~\citep{nazarenus2026actionplan}. In parallel, part-based compositional methods such as FineMoGen~\citep{zhang2023finemogen}, CoMo~\citep{huang2024controllable}, and FrankenMotion~\citep{li2026frankenmotion} target controllability at the body-part level via independent part-level prompts. These methods differ in how they handle action, direction, body-part involvement, and physical plausibility, making reliable evaluation increasingly difficult. 
% In this work, we investigate VLMs as judges of human motion generation, covering representative models from each of these lines of work.

% \paragraph{Video-language and motion benchmarks.}
% Recent video-language benchmarks show that temporal reasoning remains a weakness of general VLMs even outside the motion-capture setting. EgoSchema~\citep{mangalam2023egoschema} stresses long-form egocentric video understanding, TempCompass~\citep{liu2024tempcompass} targets event order and temporal dynamics, and Video-MME~\citep{fu2024videomme} evaluates multimodal video analysis across short and long videos. Motion-focused benchmarks narrow this gap: HumanMotionQA~\citep{endo2023motion} evaluates multiple-choice QA over motion sequences, MotionLLM introduces MoVid-Bench~\citep{chen2024motionllm} for video and motion understanding, MotionBench~\citep{hong2025motionbench} targets fine-grained video motion perception, and FAVOR-Bench~\citep{tu2025favor} adds closed- and open-ended evaluation for detailed video motion dynamics. Compared with these efforts, our benchmark emphasizes task diversity, explicit difficulty stratification, and human-solvability verification.

\paragraph{Video-language and motion benchmarks.}
Recent video-language benchmarks show that temporal reasoning remains a weakness of general VLMs even outside the motion-capture setting. EgoSchema~\citep{mangalam2023egoschema} stresses long-form egocentric video understanding, TempCompass~\citep{liu2024tempcompass} targets temporal perception across multiple aspects (e.g., action, direction, event order) and tasks, and Video-MME~\citep{fu2024videomme} evaluates multimodal video analysis across short and long videos. Motion-focused benchmarks narrow this gap: HumanMotionQA~\citep{endo2023motion} evaluates multi-step QA over motion sequences, MotionLLM introduces MoVid-Bench~\citep{chen2024motionllm} for video and motion understanding, MotionBench~\citep{hong2025motionbench} targets fine-grained video motion perception, and FAVOR-Bench~\citep{tu2025favor} adds closed- and open-ended evaluation for detailed video motion dynamics. Compared with these efforts, our benchmark emphasizes task diversity, explicit difficulty stratification, and human-solvability verification.

% \paragraph{General VLMs, judging, and embodied systems.}
% The progress of visual instruction tuning and video foundation models, including LLaVA~\citep{liu2023llava}, Video-ChatGPT~\citep{maaz2023videochatgpt}, LLaVA-OneVision~\citep{li2024llavaonevision}, Qwen2.5-VL~\citep{bai2025qwen3}, and VideoLLaMA~3~\citep{zhang2025videollama3}, motivates testing whether general-purpose VLMs can match motion-specialized systems under controlled evaluation. The same question is increasingly relevant to robotics, where vision-language-action and robot foundation models such as RT-2~\citep{brohan2023rt2}, Octo~\citep{ghosh2024octo}, OpenVLA~\citep{kim2024openvla}, $\pi_0$~\citep{black2024pi0}, and RDT~\citep{liu2024rdt} map visual observations and language instructions to physical actions. In parallel, LLM-as-a-judge and VLM-as-a-judge methods have become common for open-ended NLP and multimodal evaluation, including MT-Bench/Chatbot Arena~\citep{zheng2023judging}, G-Eval~\citep{liu2023geval}, Prometheus-Vision~\citep{lee2024prometheusvision}, and MLLM-as-a-Judge~\citep{chen2024mllmjudge}. These lines motivate our use of VLMs not only as systems under test but also as candidate judges for human motion generation.

\paragraph{General VLMs and VLM-as-a-Judge.}
The progress of visual instruction tuning and video foundation models, including LLaVA~\citep{liu2023llava}, Video-ChatGPT~\citep{maaz2023videochatgpt}, LLaVA-OneVision~\citep{li2024llavaonevision}, Qwen3.5-VL~\citep{bai2025qwen3}, VideoLLaMA~3~\citep{zhang2025videollama3}, and recent frontier systems such as the Qwen3-VL and Gemini families, motivates testing whether general-purpose VLMs can match motion-specialized systems under controlled evaluation. In parallel, LLM- and VLM-as-a-judge methods have become common for open-ended NLP and multimodal evaluation, including MT-Bench/Chatbot Arena~\citep{zheng2023judging}, G-Eval~\citep{liu2023geval}, Prometheus-Vision~\citep{lee2024prometheusvision}, MVGBench~\cite{xie2025MVGBench} and MLLM-as-a-Judge~\citep{chen2024mllmjudge}. More recently, MBench~\citep{lin2025quest} adopts VLMs as automatic judges for evaluating motion generation, but does not analyze the reliability of these judges across difficulty levels. These lines motivate our use of VLMs not only as systems under test but also as candidate judges for human motion generation, where we further stress-test their reliability from coarse to fine-grained criteria.

%% file: section/pilot_study.tex
\begin{table}[t]
\centering
\small
\resizebox{0.35\textwidth}{!}{
\begin{tabular}{lc}
\toprule
\textbf{Aggregation} & \textbf{Accuracy (\%)} \\
\midrule
Per-annotator        & 52.9 $\pm$ 4.0 \\
Pooled          & 51.8 \\
\midrule
All correct      & 34.3 \\
Either correct & 74.3 \\
\midrule
NSPose~\citep{endo2023motion} & 57.8 \\
\midrule
\multicolumn{2}{l}{\textit{Per-annotator accuracy by query type}} \\
\quad Query-Action     & 65.3 $\pm$ 12.4 \\
\quad Query-Direction  & 57.1 $\pm$ 15.7 \\
\quad Query-Body-Part  & 37.0 $\pm$ 9.9 \\
\midrule
Fleiss' $\kappa$ & 0.46 \\
\bottomrule
\end{tabular}}
\caption{Pilot study on 150 of 419 HumanMotionQA test items, each independently answered by three domain experts. Please see Appendix \ref{sec:app-pilot} for more details.}
\label{tab:pilot-results}
\end{table}

\begin{figure*}[ht]
    \centering
    \includegraphics[width=\textwidth]{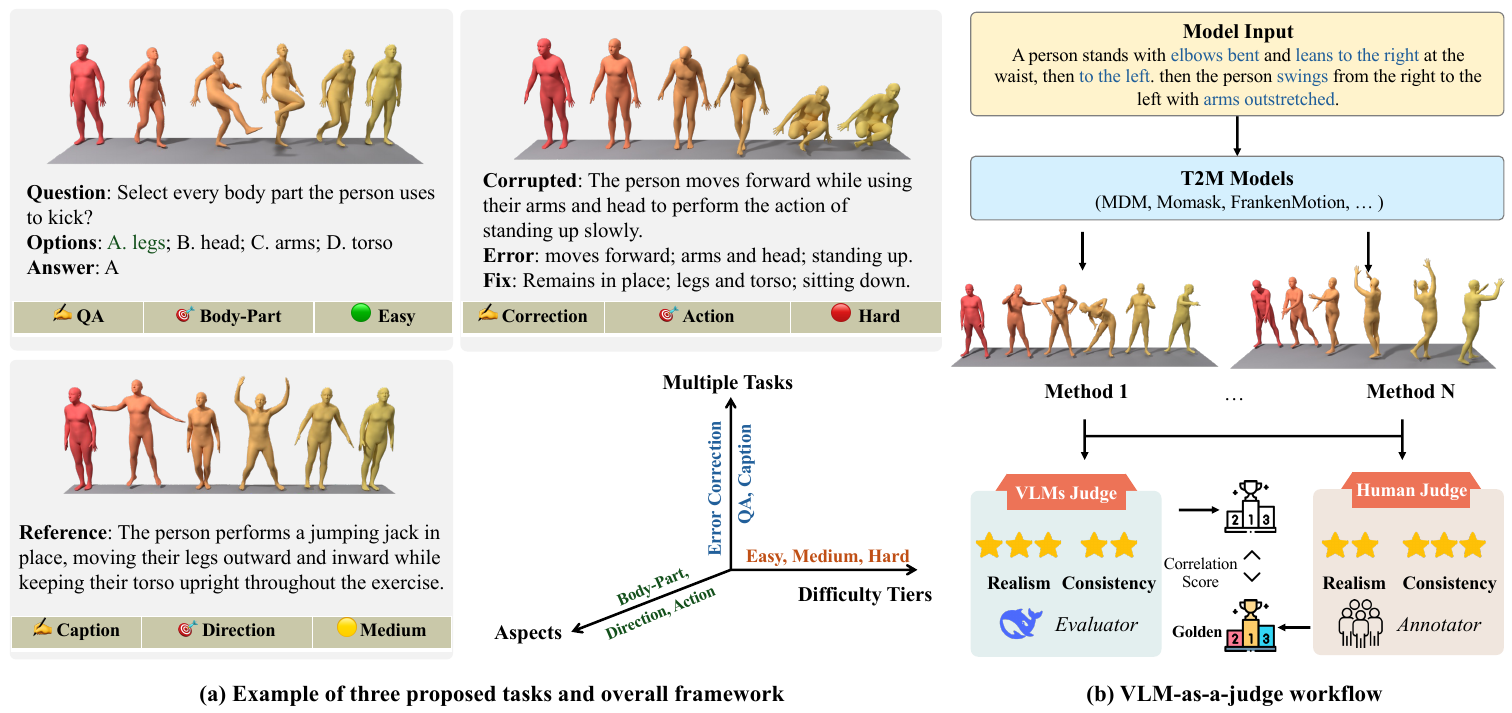}
    \caption{Dataset construction and VLM-as-a-judge evaluation workflow: (a) \benchname{} examples and design principles, where each instance is structured along $3{\times}3{\times}3$ axes including multiple tasks, semantic aspects, and difficulty tiers; and (b) Evaluation workflow for measuring alignment between human and VLM judges. 
    }
    \label{fig:overview}
\end{figure*}

\section{Pilot Study}
\label{sec:pilot}

The diagnostic value of a motion-understanding benchmark depends on whether its gold-standard answers are themselves recoverable by domain experts. When expert agreement is low, reported model accuracy becomes difficult to interpret as a measure of genuine motion understanding capability. To examine whether this condition is met in practice, we conduct a pilot study on \textsc{HumanMotionQA} \citep{endo2023motion}, a representative benchmark for motion question answering. The dataset is built on motion sequences from \textsc{Amass} \citep{AMASS} together with the rule-based QA templates of \textsc{Babel-QA} \citep{endo2023motion}, and is broadly representative of the QA formulation adopted in subsequent work. We recruited three motion-domain experts to independently answer a uniformly random sample of test items through a custom blind web interface; setup details, annotator details, and website interface are provided in Appendix~\ref{sec:app-pilot-setup}.

% \XH{shouldn't we emphasize here the inconsistency of its annotation? Benchmark being difficult is not a problem, but inconsistent is a real problem}

The results as shown in Table~\ref{tab:pilot-results}, reveal severe annotation inconsistency in \textsc{HumanMotionQA} that fundamentally limits its solvability. Per-annotator accuracy averages $52.9\%$ and unanimous agreement reaches $34.3\%$, both of which are comparable to the $57.8\%$ accuracy reported for the motion-specific NSPose model on
the same split. Inter-annotator agreement (IAA) is moderate (Fleiss' $\kappa = 0.46$), suggesting that a non-trivial portion of the difficulty stems from how items are framed rather than from annotator-specific variation. 

A manual audit of items on which experts disagreed with the gold labels (Appendix~\ref{sec:app-failure-taxonomy})
reveals four recurring patterns in the source annotations:
{(1)} body-part granularity collisions, where
hierarchical labels such as hand and arm can be hard to disentangle; {(2)} ambiguous spatial frames of reference in directional queries; {(3)} temporal scope mismatches in sequential questions; and {(4)} composite-action dominance, when multiple simultaneous actions are present. These patterns are useful design signals for \benchname{}: we encode each as an explicit rejection criterion in our expert-filtering rubric (\S\ref{sec:construction}), so that retained items can be answered from the visible motion alone.

% \begin{table}[t]
% \centering
% \small
% \begin{tabular}{lc}
% \toprule
% \textbf{Aggregation} & \textbf{Accuracy (\%)} \\
% \midrule
% Per-annotator (mean $\pm$ std) & 51.1 $\pm$ 5.1 \\
% Majority (2 of 3)              & 53.3 \\
% Unanimous (3 of 3)             & 30.0 \\
% \midrule
% NSPose~\citep{endo2023motion} (model) & 57.8 \\
% \midrule
% \multicolumn{2}{l}{\textit{Per-annotator accuracy by query type}} \\
% \quad Query-Action     & 33.3 $
% \pm$ 15.3  \\
% \quad Query-Direction  & 86.7 $\pm$ 5.8 \\
% \quad Query-Body-Part  & 33.3 $\pm$ 11.5 \\
% \midrule
% Fleiss' $\kappa$       & 0.44 \\
% \bottomrule
% \end{tabular}
% \caption{Pilot study results on randomly sampled HumanMotionQA test items, independently answered by three domain experts. Expert accuracy falls markedly below the 57.8\% reported for the NSPose model.}
% \label{tab:pilot-results}
% \end{table}

%% file: section/benchmark.tex
\section{\benchname{} Benchmark}
\label{sec:construction}

In this section, we first introduce \benchname{} along three parts: task design, construction pipeline and principles, and quality control; and then discuss the VLM-as-a-judge protocol.

\begin{table}[t]
\centering
\small
\setlength{\tabcolsep}{4pt}
\resizebox{\columnwidth}{!}{
\begin{tabular}{l c c c c}
\toprule
\textbf{Benchmark}  & \textbf{\#Ex.} & \textbf{Multi-task} & \textbf{Multi-Axes} & \textbf{Diff.} \\
\midrule
BABEL-QA \cite{BABEL:CVPR:2021}                      & 2.6k     & \xmark & \cmark & \xmark \\
HuMMan-QA  \cite{li2025imore}                    & 3.1k     & \xmark & \cmark & \xmark \\
MoVid-Bench \cite{chen2025motionllm}                  & 533 & \pmark & \pmark & \xmark \\
MotionBench  \cite{hong2025motionbench}                  & 4.0k     & \xmark & \pmark & \xmark \\
FineBench    \cite{faure2026finebench}                & 199k     & \xmark & \pmark & \xmark \\
HMI-Bench    \cite{song2025towards}                   & 115k     & \xmark & \pmark & \xmark \\
KPM-Bench   \cite{lin2026kpm}                   & 75k      & \cmark & \cmark & \xmark \\
\midrule
\rowcolor{gray!15}
\textbf{NextMotionQA (Ours)} & \textbf{1.3k} & \cmark & \cmark & \cmark \\
\bottomrule
\end{tabular}}
\caption{Comparison of \benchname{} with existing human-motion benchmarks. \cmark / \pmark / \xmark\quad denotes full / partial / no support. Multi-task means all three formats; Multi-Axes: all three semantic axes; Diff. means explicit difficulty stratification.}
\label{tab:benchmark_comparison}
\end{table}

\paragraph{Formalization.}
Each instance in \benchname{} is a tuple attribution:
\begin{equation}
(m_i,\, t_i,\, a_i,\, d_i,\, y_i),\quad i=1,\dots,N,
\label{eq:benchmark}
\end{equation}
where $m_i$ is a 3D motion clip drawn from AMASS, $t_i\!\in\!\{T_1,T_2,T_3\}$ is the task format, $a_i\!\in\!\{A_1,A_2,A_3\}$ the semantic axis, $d_i\!\in\!\{\text{Easy},\text{Medium},\text{Hard}\}$ the difficulty tier, and $y_i$ the gold answer\footnote{$y_i$ is not applicable to $T_2$. Depending on Task requirement, SMPL data can also be collected.}. As shown in Figure \ref{fig:overview}(a), the factorization $(t, a, d)$ is itself the contribution: stratifying by task, axis, and difficulty localizes where a model fails rather than merely whether it does, providing the diagnostic resolution that the pilot (\S\ref{sec:pilot}) showed prior benchmarks to lack.

\begin{table}[t]
\centering
\small
\setlength{\tabcolsep}{4pt}
\resizebox{0.49\textwidth}{!}{
\begin{tabular}{l|ccc}
\toprule
Statistic & \textbf{T1: QA} & \textbf{T2: Caption} & \textbf{T3: Correction} \\
\midrule
Input & \{$m_i$, $q_i$, $y_i$\}. & \{$m_i$\} & \{$m_i$, $c_i$\} \\
Output & \{$y_i$\} & \{$c_i$\} & \{$e_i$, $c_i$\} \\
\midrule
\# examples            & 511 & 396 & 400 \\
\# Videos        & 483 & 396 & 400 \\
% \# options             & 4   & --  & --  \\
\midrule
% \multicolumn{4}{l}{\textit{By axis}}\\
A1 body-part     & 172 & 132 & 136 \\
A2 direction     & 163 & 132 & \phantom{0}92 \\
A3 action        & 176 & 132 & 172 \\
\midrule
% \multicolumn{4}{l}{\textit{By difficulty (\# / avg.\ dur.\ s)}}\\
Easy             & \phantom{0}7.6  & \phantom{0}8.2  & \phantom{0}8.0 \\
Medium           & \phantom{0}9.4  & \phantom{0}9.8  & \phantom{0}9.3 \\
Hard             & 12.4 & 15.3 & 13.5 \\
% \midrule
% % \multicolumn{4}{l}{\textit{Content}}\\
% Text Len.  & 12.8 & 23.3 & 21.7 \\
% Targets / ex.          & 1.11 & 2.58 & 2.12 \\
\bottomrule
\end{tabular}}
\caption{\textbf{NextMotionQA statistics.} Three semantic axes (A1 body-part,
A2 direction, A3 action) $\times$ three difficulty tiers. 992 unique SMPL-H
clips (30\,fps) from 16 AMASS subsets; metadata from BABEL + HumanML3D.
``avg.\ dur.'' is mean clip length in seconds. T3 axis counts are by primary
axis (errors may span multiple axes) and are not balanced to a $3\times3$ grid.  }
% \XH{T1: why less videos than examples?}} Because some cases use the same videos.
\label{tab:stats}
\end{table}

\paragraph{Task design: recognize, describe, critique.}
The three formats are designed to probe progressively stronger capabilities rather than to evaluate independent skills in parallel. $T_1$ (multi-select QA over four options) probes {recognition}: given a closed candidate set, can the model identify the licensed subset? Multi-select is essential, since single-choice MCQ lets a model trade specificity for safety by defaulting to the most generic option, masking the granularity errors this benchmark targets. $T_2$ (free-form captioning) probes {description}, strictly extending $T_1$ by requiring production rather than selection: can the model spontaneously surface the axis-relevant attribute in open vocabulary? $T_3$ (caption error correction) probes {critique}, and is strictly the hardest of the three because it composes recognition (which span is wrong), description (what the span should be), and rejection of a fluent distractor an under-grounded model would otherwise accept. This ladder also mirrors downstream T2M evaluation, where the practical question is not whether the motion matches the prompt but where it fails and how to fix it, which is precisely the use case our VLM-as-a-judge experiments (\S\ref{sec:judge}) target.

\paragraph{Semantic axes.}
Motivated by \citet{li2026frankenmotion}, orthogonal to task format, every item is tagged with one of three load-bearing axes of motion understanding identified in the pilot: $A_1$ body-part involvement ({which} parts execute the motion), $A_2$ translation direction ({which way} the body moves), and $A_3$ action semantics ({what} the motion is). Each axis is operationalized to avoid the failure modes F1--F4 (\S\ref{sec:app-failure-taxonomy}).

\paragraph{Difficulty.}
Difficulty is assigned at the clip level, so a clip carries the same $d$ across all three task formats, via:
\begin{equation}
d(m) =
\begin{cases}
\text{Easy}   & \!\!\text{if } |\mathcal{L}(m)|\!=\!1 \,\wedge\, \mathcal{M}(m)\!=\!\emptyset, \\
\text{Medium} & \!\!\text{if } |\mathcal{L}(m)|\!=\!2, \\
\text{Hard}   & \!\!\text{if } |\mathcal{L}(m)|\!\geq\!3 \,\vee\, \mathcal{M}(m)\!\neq\!\emptyset,
\end{cases}
\label{eq:difficulty}
\end{equation}
where $\mathcal{L}(m)$ is the set of BABEL action labels overlapping $m$ and $\mathcal{M}(m)$ the set of compositional modifiers (direction, speed, manner, fine-grained body part) extracted from the HumanML3D caption. Tying difficulty to the clip rather than the question is what makes cross-task comparison meaningful: a model's $T_2 \to T_3$ degradation on the same $m$ is attributable to task format, not to a confounded change in motion complexity.

\subsection{Construction Pipeline}
\label{sec:pipeline}
\benchname{} is built by a two-pass semi-automatic pipeline in which a single VLM, Qwen3.6-Plus, is queried twice with different inputs, followed by hard quota allocation and domain expert verification.

\paragraph{Pass 1: metadata-conditioned drafting.}
The VLM receives only the BABEL action labels and HumanML3D caption for clip $m$, not the rendered video, and drafts a candidate item per (task, axis) cell following a cell-specific template (Appendix~\ref{ax:all_prompt_settings}): a $T_1$ option set with multi-select gold, a $T_2$ reference caption, or a $T_3$ corrupted caption with labelled span and rewrite. Withholding the video grounds the draft in human-curated symbolic labels rather than the VLM's own perception.

\paragraph{Pass 2: video-conditioned refinement.}
The same VLM is then queried with both the rendered video and the Pass-1 draft, and revises any content inconsistent with the visible motion. This corrects the residual gap where metadata is technically correct but mismatched to what the video shows.

\paragraph{Quota allocation.}
The pipeline targets a $|T|\times|A|\times|D|=27$-cell grid with a hard per-cell quota: over-generation is rejection-sampled down, and under-generated cells trigger re-drafting. Quota is enforced before expert verification (\S\ref{sec:scoring}) to keep distributional control separate from quality control. 
% The final composition ($1{,}336$ items over $992$ clips; $540/396/400$ for $T_1/T_2/T_3$, Table~\ref{tab:stats}) reflects $T_1$'s denser multi-select sampling and the primary-axis categorization of $T_3$.

\paragraph{Scalability.}
Both passes use a single VLM under fixed prompts, so scaling the benchmark requires only more source clips and verifier hours, not pipeline redesign.

\begin{table*}[ht]
\centering
\small
\setlength{\tabcolsep}{3pt}
\resizebox{0.97\textwidth}{!}{
\begin{tabular}{l|ccccccccccc|c}
\toprule
\multirow{2}{*}{\textbf{Model}}
 & \multicolumn{3}{c}{\textbf{Task 1}}
 & \multicolumn{4}{c}{\textbf{Task 2}}
 & \multicolumn{4}{c|}{\textbf{Task 3}}
 & \multirow{2}{*}{\textbf{Avg.}} \\
\cmidrule(lr){2-4} \cmidrule(lr){5-8} \cmidrule(lr){9-12}
 & Acc. & Jaccard & Prec. & Easy & Med. & Hard & Avg & Identify & Recall & Correct & Avg & \\
\midrule
\multicolumn{13}{c}{\textit{Open Source Model}} \\
\midrule
Qwen3.5-0.8B          & 9.77  & 34.24 & 34.92 & 31.00 & 16.60 & 22.30 & 23.30 & 12.91 & 14.69 & 6.62  & 11.41 & 20.34 \\
Qwen3.5-4B            & 18.80 & 44.19 & 44.55 & 42.70 & 31.90 & 30.00 & 34.87 & 64.18 & 69.30 & 35.57 & 56.35 & 42.36 \\
Qwen3.5-9B            & 22.93 & 47.95 & 48.93 & 39.40 & 31.70 & \underline{30.10} & 33.73 & 60.51 & 64.37 & 32.15 & 52.34 & 42.00 \\
Qwen3.5-27B           & 40.23 & 55.78 & 56.45 & 43.30 & \underline{36.10} & 27.30 & \underline{35.57} & 70.59 & 78.08 & \underline{39.87} & 62.85 & 49.75 \\ \midrule
InternVL3.5-4B        & 26.50 & 47.43 & 48.07 & 35.40 & 29.70 & 29.50 & 31.53 & 41.48 & 44.30 & 18.95 & 34.91 & 35.70 \\
InternVL3.5-8B        & 29.51 & 48.17 & 49.03 & 39.40 & 29.10 & 27.60 & 32.03 & 57.09 & 62.13 & 34.35 & 51.19 & 41.82 \\
InternVL3.5-14B       & 3.76  & 37.99 & 38.74 & \underline{44.60} & 30.10 & 27.80 & 34.17 & 43.67 & 46.32 & 31.65 & 40.55 & 33.85 \\ \midrule
LLaVA-1.5-4B          & 43.42 & 51.32 & 53.92 & 37.70 & 23.80 & 26.70 & 29.40 & 55.19 & 59.90 & 23.38 & 46.16 & 41.70 \\
LLaVA-1.5-8B          & 42.86 & 52.41 & 54.15 & 35.80 & 25.20 & 25.40 & 28.80 & 54.26 & 60.56 & 27.30 & 47.37 & 42.00 \\
\midrule
\multicolumn{13}{c}{\textit{Close Source Model}} \\
\midrule
GPT-5.4-mini$^{\dagger}$       & 44.74 & 52.27 & 54.97 & 40.60 & 28.10 & 29.60 & 32.77 & 54.01 & 58.25 & 18.44 & 43.57 & 42.33 \\
Qwen3.6-Plus$^{\dagger}$       & \textbf{60.15} & \textbf{64.52} & \underline{67.51} & 44.40 & 32.00 & 29.30 & 35.23 & \underline{74.50} & \underline{82.84} & 38.40 & \underline{65.25} & \underline{54.85} \\
Gemini-3.1-Flash$^{\dagger}$   & \underline{59.02} & \underline{64.07} & \textbf{67.58} & \textbf{52.30} & \textbf{42.40} & \textbf{35.90} & \textbf{43.53} & \textbf{76.79} & \textbf{82.95} & \textbf{44.94} & \textbf{68.23} & \textbf{58.44} \\
\bottomrule
\end{tabular}}
\caption{Main results across three tasks proposed in \benchname{}. 
For Task 1 we report Accuracy (Acc), Jaccard, and Precision; for Task 2 we report results on Easy / Medium / Hard subsets and their average (Avg); for Task 3 we report Identify, Token Recall, Correct, and their average (Avg). 
Avg. in the last column denotes the macro-average across the three task-level averages.
The \textbf{best} result in each column is in bold and the \underline{second-best} is underlined.}
\label{tab:main_results}
\end{table*}

\subsection{Quality Control}
\label{sec:scoring}

We guarantee annotation quality through three aspects: source provenance, automatic VLM refinement, and expert verification.

\paragraph{Stage 1: source provenance.}
Every clip is drawn from AMASS with metadata from BABEL and HumanML3D, all three being peer-reviewed, expert-curated resources whose label quality has been validated at corpus scale. \benchname{} inherits this provenance: an item is admissible only if its underlying source labels survive the pilot's F1--F4 audit (See Appendix~\ref{sec:app-failure-taxonomy} for more details).

\paragraph{Stage 2: automatic VLM refinement.}
Pass-2 of the construction pipeline (\S\ref{sec:pipeline}) doubles as a quality filter: drafts whose video-conditioned revision diverges from the metadata-conditioned draft are flagged for expert review rather than auto-accepted. Items that pass this stage are video-consistent under the same VLM that drafted them, eliminating the easy class of metadata-only errors.

\paragraph{Stage 3: expert verification.}
Every filtered item is independently judged by three motion-domain experts as \texttt{accept}, \texttt{revise}, or \texttt{reject}, with rejection mandatory on any F1--F4 violation. Acceptance requires unanimous \texttt{accept} from all three experts; \texttt{revise}-flagged items are jointly edited and re-judged, and \texttt{reject}-flagged items are discarded outright. 

% Inter-annotator agreement on the accept/reject decision is Cohen's $\kappa\!=\!\fillme{0.XX}$, and the funnel produces $1{,}336$ accepted items from \fillme{$N_\text{draft}$} drafts (yield \fillme{XX}\%).

% The three stages are deliberately structured so that the pipeline degrades gracefully under scale: stages 1 and 2 are fully automatic, and only stage 3 is human-bottlenecked. Practitioners wishing to extend \benchname{} can easily apply stages 1--2 to fresh AMASS sub-corpora, deferring expert hours to the residual that stage 2 cannot resolve.
 \paragraph{Comparison with existing Datasets.} 
As shown in Table~\ref{tab:benchmark_comparison} and \ref{tab:stats}, prior benchmarks typically focus on isolated tasks or limited evaluation axes, while \textbf{NextMotionQA} provides the first unified benchmark featuring multi-task evaluation, multi-axis reasoning, and difficulty-aware assessment.

\subsection{VLM-as-a-Judge Protocol}
\label{sec:judge}

\paragraph{Setup.} In a complementary direction, recent work has begun using VLMs as judges for T2M evaluation, where feature-space metrics (FID, MM-Dist, Diversity) correlate weakly with human perception. Given the capability gaps \benchname{} reveals on motion understanding, we ask whether VLM judges exhibit the same degradation under harder tasks. As shown in Figure~\ref{fig:overview}(b), we reproduce representative T2M systems, render their outputs alongside ground-truth AMASS motions, and elicit ratings from both human experts (gold) and the VLM judge on two Likert criteria: {realism}, whether the motion resembles plausible human movement, and {semantic consistency}, whether it matches the prompt. We deliberately restrict the judge to a base prompting setup without multi-agent ensembling or post-training, since our goal is to probe the model's intrinsic capability boundary rather than to optimize judge performance. 

\paragraph{Human Comparsion.} Within this unified setup, we apply the same protocol under three version of escalating complexity: {V1} single-action prompts on small-scale motion clips, where judgment reduces to near-recognition; {V2} composite, chained actions on larger-scale clips, which require tracking temporal composition; and {V3} fine-grained, temporally segmented motions, which jointly resolve body-part ($A_1$) and temporal index, the capability \S\ref{sec:exp-main} flags as weakest in current VLMs. Across all three versions, human preferences serve as the gold reference for alignment measurement.
% , and judge--expert alignment is measured by correlation at instance, per-question, and system granularity.

% \paragraph{Statistics.}
% Final dataset composition is summarized in Table~\ref{tab:stats}; extended quality controls (caption type-token ratio versus HumanMotionQA, sanity-check expert accuracy on a random subsample) are in Appendix~\ref{app:data}.

%% file: section/experiment.tex
\section{\benchname{} Setup}
\label{sec:exp-setup}
  
%%From Chuqiao Baseline desctiption
% \paragraph{Baselines.}
% We evaluate VLMs as judges on nine representative text-to-motion methods spanning four lines of work. From the diffusion family, we include MDM~\citep{tevet2023mdm}, which establishes continuous motion synthesis, and MotionLCM~\citep{dai2024motionlcm}, a consistency-model variant that improves sampling efficiency. From the discrete-token family, we include MoMask~\citep{guo2024momask}, which casts motion as a language-like sequence. From the continuous-latent family, we include MotionStreamer~\citep{xiao2025motionstreamer}, which introduces continuous causal latents with autoregressive diffusion; MARDM~\citep{meng2024rethinking}, which performs bidirectional masked autoregression in continuous space; and ActionPlan~\citep{coral2025actionplan}, which uses per-frame text-latent anchors with latent-specific diffusion timesteps. From the part-based family, we include FineMoGen~\citep{zhang2023finemogen}, CoMo~\citep{huang2024como}, and FrankenMotion~\citep{li2026frankenmotion}, which decompose motion into per-part units guided by independent part-level textual prompts.

\paragraph{Evaluated VLMs.}
We benchmark twelve VLMs spanning including: {(i)~Open-source VLMs}: Qwen3.5-VL at four scales (0.8B, 4B, 9B, 27B)~\citep{bai2025qwen3}, InternVL3.5 at three scales (4B, 8B, 14B), and LLaVA-1.5 at two scales (4B, 8B); and {(ii)~Frontier closed-source VLMs}: GPT-5.4-mini, Qwen3.6-Plus, and Gemini-3.1-Flash, representing the current upper boundary for human motion understanding. Full model versions and access dates are in Appendix~\ref{ax:evaluated_models}. We also provide all prompts in Appendix~\ref{ax:all_prompt_settings} to ensure reproducibility.

\paragraph{VLM-as-A-Judge Method Selection.}
For Version 1, we use 20 videos from MARDM~\citep{meng2025rethinking}, MotionStreamer~\citep{xiao2025motionstreamer} and ActionPlan~\citep{nazarenus2026actionplan}. For Version 2 with 50 videos, we adopt three methods namely MDM~\citep{tevet2023mdm}, MotionLCM~\citep{dai2024motionlcm} and MoMask~\citep{guo2024momask} for comparison. And for Version 3 with 20 videos, we adopt FineMoGen~\citep{zhang2023finemogen}, CoMo~\citep{huang2024controllable}, and FrankenMotion~\citep{li2026frankenmotion}. Method details please refer to Appendix~\ref{ax:vlm_judges_baseline}.

\paragraph{Metrics.}
For \textbf{$T_1$}, we report exact-match Accuracy, Jaccard, and Precision over the predicted and gold option subsets, with Jaccard as the headline metric since exact-match under-credits the partial overlap multi-select routinely produces. For \textbf{$T_2$}, a Qwen3.6-Plus judge scores each caption on content coverage and action consistency (both $[0,100]$); we report the mean within each difficulty subset (Easy/Med/Hard) and overall. For \textbf{$T_3$}, we report Identify (proportion of gold error spans recovered), Recall (token-level recall against the gold span), and Correct (Gemini-judged semantic correctness of the rewrite, allowing synonyms); the task-level score is their average.
% The headline overall score averages $T_1$ Jaccard, $T_2$ Avg, and $T_3$ Avg.

\begin{figure}[t]
    \centering
    \includegraphics[width=0.48\textwidth]{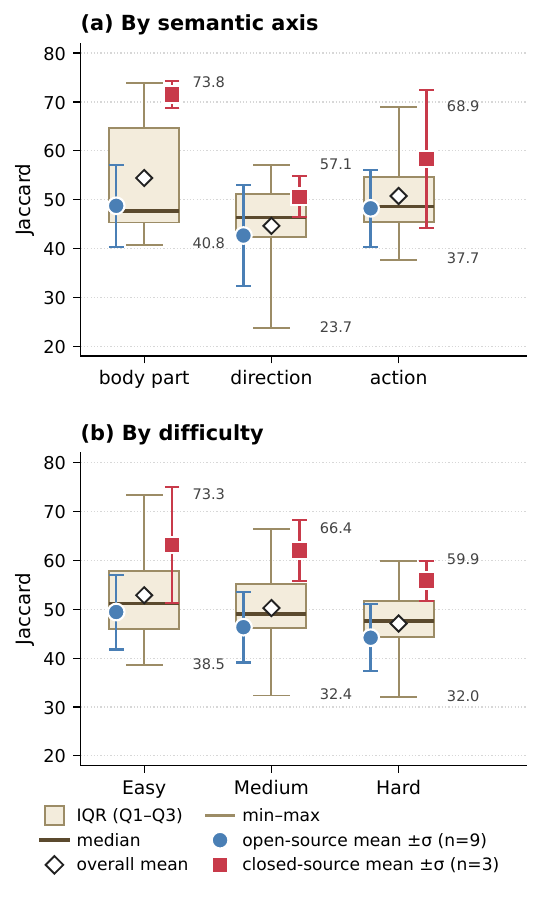}
    \caption{%
        Task 1 (MQA) Jaccard breakdown. (a) By semantic axis: nearly all models
        show a V-shape, with direction being the universally hardest
        sub-axis. (b) By difficulty: top-tier models exhibit a clear monotonic decline from Easy to Hard,
        whereas weaker models remain flat and they fail even on easy questions.
    }
    \label{fig:t1_breakdown}
\end{figure}
  
\section{Evaluation and Analysis}
\label{sec:exp-main}
Table~\ref{tab:main_results} reports extensive results for the twelve VLMs, supporting the following observations.

\subsection{Capability Analysis}

\begin{figure}[t]
    \centering
    \includegraphics[width=0.49\textwidth]{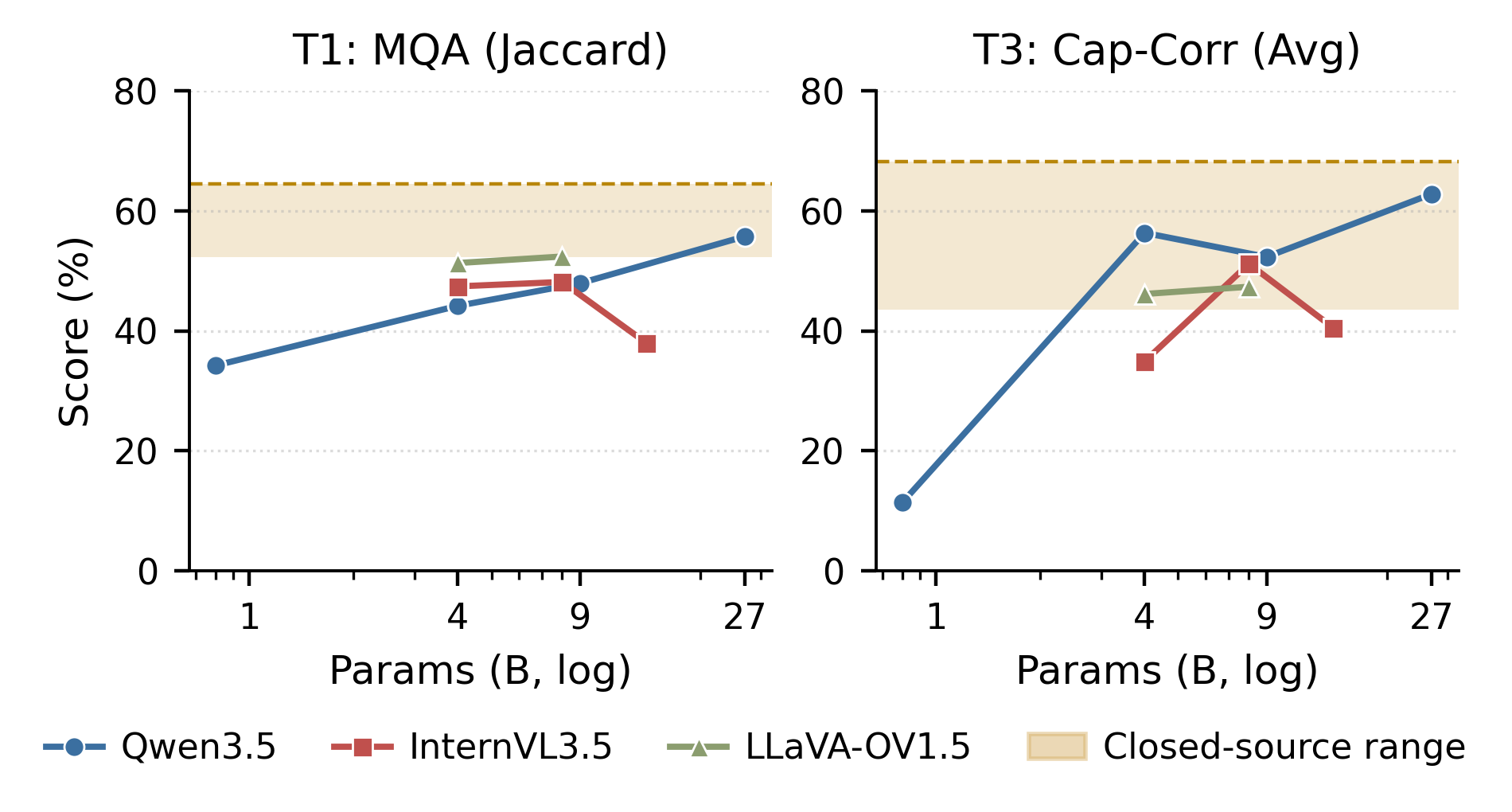}
    \caption{Scaling behavior shows mixed effects: Qwen3.5 moves toward the frontier band, while InternVL3.5 regresses at 14B. T3 performance saturates beyond 4B, with Qwen3.5-27B entering the band.}
    \label{fig:scaling}
  \end{figure}

\paragraph{Open vs.\ Closed VLMs.}
Closed-source frontier systems lead the leaderboard, with Gemini-3.1-Flash achieving the highest overall score (58.44) and Qwen3.6-Plus second (54.85). The strongest open-source system, Qwen3.5-27B, reaches 49.75, leaving a 8.69-point gap to the closed-source ceiling. The gap is not uniform across the closed-source tier, however: GPT-5.4-mini scores 42.33 overall and trails several mid-scale open-source models on $T_3$ correction (43.57 vs.\ 62.85 for Qwen3.5-27B), indicating that the closed--open boundary tracks training-mix quality rather than the open/closed label itself. Within each open-source family, scaling typically helps (Qwen3.5: 20.34$\to$49.75 from 0.8B to 27B) also as shown in Figure \ref{fig:scaling}, but not monotonically: InternVL3.5-14B underperforms both its 4B and 8B siblings on $T_1$ Accuracy (3.76\footnote{This abnormally low score is primarily attributed to InternVL3.5-14B’s tendency to generate multiple answers, which significantly reduces evaluation accuracy.} vs.\ 26.50 / 29.51), illustrating that parameter count alone does not guarantee motion-understanding competence within a family.

\paragraph{Task difference.}
$T_2$ captioning is the universal bottleneck across all twelve VLMs: the top model reaches only 43.53 on $T_2$ versus 64.07 on $T_1$ Jaccard and 68.23 on $T_3$, and no open-source system exceeds 35.57 on $T_2$ despite several exceeding 50 on $T_1$. This pattern validates the recognize, describe, critique posited in \S\ref{sec:pipeline}: open-vocabulary production of axis-relevant attributes is strictly harder than closed-set selection or local span editing. Within $T_1$, Jaccard sits far above exact-match Accuracy across nearly every model, exposing a specificity vs.\ coverage trade-off in which models identify {some} but not {all} options, the granularity error a single-choice MCQ would mask.

\paragraph{Aspects and difficulty tiers.}
Figure~\ref{fig:t1_breakdown} decomposes $T_1$ Jaccard along two design axes and reveals two robust patterns. Direction ($A_2$) is the universal weak sub-axis: every family shows a V-shape, and the closed vs.\ open gap narrows from 25.0 points on $A_1$ (73.8 vs.\ 48.8) to only 8.2 points on $A_2$ (50.7 vs.\ 42.5), capping both families near a common ceiling and suggesting a shared pretraining blind spot in camera-frame temporal grounding rather than a capacity issue. Difficulty stratification additionally separates model tiers: closed-source models exhibit a monotonic Easy to Hard decline (73.3 to 59.9) that tracks the compositional complexity our rubric encodes, whereas open-source models remain flat or invert, showing they are bottlenecked on capabilities already required by Easy items.

\subsection{VLM-as-A-Judge for motion generation}

\paragraph{Overall performance.}
Table~\ref{tab:alignment} reports judge vs.\ human agreement across V1 (single-action), V2 (composite-action), and V3 (fine-grained part-level) with Gemini-3.1-Flash as judge, tracing a clean V1 to V3 degradation curve. V1 shows close alignment at every granularity (instance $r\!=\!0.774$, $\kappa\!=\!0.701$, system $r\!=\!0.966$); V2 halves instance and per-question scores (0.495 / 0.346) yet preserves system-level ranking ($r\!=\!1.000$), an instance vs.\ system dissociation that keeps the judge usable for model comparison but not per-clip analysis; V3 collapses across all granularities and the system-level Pearson flips sign ($r\!=\!-0.146$), leaving the judge anti-correlated with human preference at the part-level motion editing most needs.

\begin{table}[t]
  \centering
  \small
  \resizebox{0.46\textwidth}{!}{
  \begin{tabular}{llccc}
  \toprule
  \textbf{Granularity} & \textbf{Metric} & \textbf{V1} & \textbf{V2} & \textbf{V3} \\
  \midrule
  \multirow{2}{*}{Instance}     & Pearson $r$      & \textbf{0.774} & 0.495 & 0.116 \\
                                & Spearman $\rho$  & \textbf{0.737} & 0.485 & 0.071 \\
  \midrule
  \multirow{2}{*}{Per-question} & Accuracy         & \textbf{0.800} & 0.560 & 0.400 \\
                                & Cohen's $\kappa$ & \textbf{0.701} & 0.346 & 0.104 \\
  \midrule
  System                        & Pearson $r$      & 0.966 & \textbf{1.000} & $-0.146$ \\
  \midrule
  \multicolumn{2}{l}{\# clips}                     & 20 & 50 & 20 \\
  \bottomrule
  \end{tabular}}
  \caption{VLM--human agreement at three granularities, ordered by decreasing
  overall alignment. Instance and per-question levels score each clip; the system
  level correlates aggregate per-method preference. V1/V2 are text-to-motion;
  V3 is part-level conditioned motion.}
  \label{tab:alignment}
  \end{table}
  
\paragraph{Why does VLM-as-a-judge fail under fine-grained motion?}
The V3 collapse is not an isolated judge failure, but a direct manifestation of the capability gaps identified in \S\ref{sec:exp-main}. Reliable V3 judgment requires resolving body-part identity ($A_1$) and temporal localization jointly, the same joint capability our understanding evaluation flagged as weakest in current VLMs along two complementary dimensions: temporal grounding under-performs across all families on the direction axis ($A_2$), and captioning fluency on $T_2$ does not transfer to localizing the same attribute when it appears in error ($T_3$). The V1$\to$V3 trajectory therefore delineates a clear operating envelope: VLM-as-a-judge is reliable for coarse criteria and system-level ranking, but not yet suitable for fine-grained part-level evaluation. We identify camera-frame grounding, temporal localization, and part-aware pretraining as the joint prerequisites for closing this gap.

%% file: section/conclusion.tex
\section{Conclusion}

In this paper, we introduce \benchname{}, a $3\!\times\!3\!\times\!3$ benchmark of 1,307 expert-verified instances that decomposes 3D human motion understanding along three task formats, three semantic axes, and three difficulty tiers. Based on it, benchmarking twelve representative VLMs exposes structural gaps a single aggregate score would hide: no system dominates across the three formats, translation direction is the universal weak axis. Building on this observation, we further demonstrate that a strong VLM judge tracks human preference closely on coarse T2M criteria (Cohen's $\kappa\!=\!0.70$) but collapses on fine-grained part-level judgment ($\kappa\!=\!0.10$), validating VLM-as-a-judge in its strong regime while clarifying where camera-frame, temporal, and part-aware pretraining remain the prerequisites for fine-grained motion evaluation.

\paragraph{Limitations.}
Overall, we believe there are three limitations bound the conclusions above:
(i) All motion data are drawn from AMASS and inherit its distribution; out-of-domain motions such as dyadic interaction, hand-object manipulation, or non-rigid deformation lie outside the present scope. (ii) The VLM-as-a-judge protocol uses a single fixed standard, which under-resolves back-of-body configurations relevant to part-level plausibility; multi-view rendering is a clear next step. (iii) Expert verification, while necessary for label quality, caps the benchmark at the order of $10^3$ items; scaling to $10^4$ will require an expert-in-the-loop active-learning regime that we leave to future work.

\paragraph{LLMs Usage.} Through the paper, we use LLMs to assist with grammar checking and minor rephrasing for clarity. LLMs did not contribute to the conceptual design of the study, experimental implementation, or core writing of the paper.

%% file: appendix.tex
% \clearpage
\appendix

\section{Pilot Study Supplementary Material}
\label{sec:app-pilot}

This appendix provides extended details regarding the experimental setup, participant background, statistical analyses, and qualitative failure cases identified in the pilot study discussed in \S\ref{sec:pilot}.

\subsection{Detailed Setup and Demographics}
\label{sec:app-pilot-setup}

\paragraph{Data Sampling and Rendering.}
The 150 sampled items were drawn uniformly from the BABEL-QA test split of HumanMotionQA without stratification to preserve the natural distribution of the source dataset. This sample consists of 419 queries, 80 direction queries, 275 action queries, and 64 body-part queries. The underlying raw motion coordinates were rendered using Blender into standardized video clips at 30 FPS with a resolution of $768 \times 768$ pixels. This rendering pipeline ensures that the visual quality, camera angle, and framerate are identical across all evaluated items, eliminating any low-level visual confounding factors.

\paragraph{Expert Demographics.}
Our three domain experts are active researchers in computer vision and computer graphics, specializing in 3D human motion modeling, deep generative models, and character animation. Each expert holds at least a Master’s degree (with three holding or pursuing a PhD) and has published peer-reviewed papers in top-tier vision or graphics venues. This guarantees a level of domain expertise, spatial reasoning, and attention to detail that far exceeds that of the crowdworkers who originally annotated the BABEL dataset.

\paragraph{Annotation Interface.}
As shown in Figure~\ref{fig:human-study-interface}, we deployed a custom, web-based annotation interface to conduct the study. Key design features of this interface include:
\begin{itemize}
    \item \textbf{Blind Evaluation}: To prevent band-wagoning or herd behavior, experts had no access to the gold labels or the choices made by other participants.
    \item \textbf{Arbitrary Replay}: Experts could loop, pause, or scrub through the rendered motion videos indefinitely before submitting an answer.
    \item \textbf{Fine-grained Confidence Rating}: For every item, experts were forced to report their confidence on a 5-point Likert scale (1: Low confidence/guessing, 5: High confidence/certainty) alongside their selected choice.
\end{itemize}

\subsection{Expert Confidence Analysis}
\label{sec:app-confidence}
We performed a correlation analysis to investigate whether experts could self-calibrate when facing highly ambiguous labels. Our findings indicate a strong presence of the {``confidently wrong''} phenomenon. Experts reported a high confidence score ($\ge 4/5$) on 69.2\% of all items. However, their empirical accuracy on these high-confidence responses was only 65.5\%. 

Furthermore, the Pearson correlation coefficient between expert confidence and correctness is $r = 0.272$. This extremely weak correlation demonstrates that experts do not merely guess when faced with uncertainty; instead, they commit to specific, logically defensible answers that nevertheless diverge from the gold labels due to hidden annotation conventions in the original dataset.

\subsection{Complete Failure Taxonomy}
\label{sec:app-failure-taxonomy}
\begin{figure}[t]
\centering
\includegraphics[width=\linewidth]{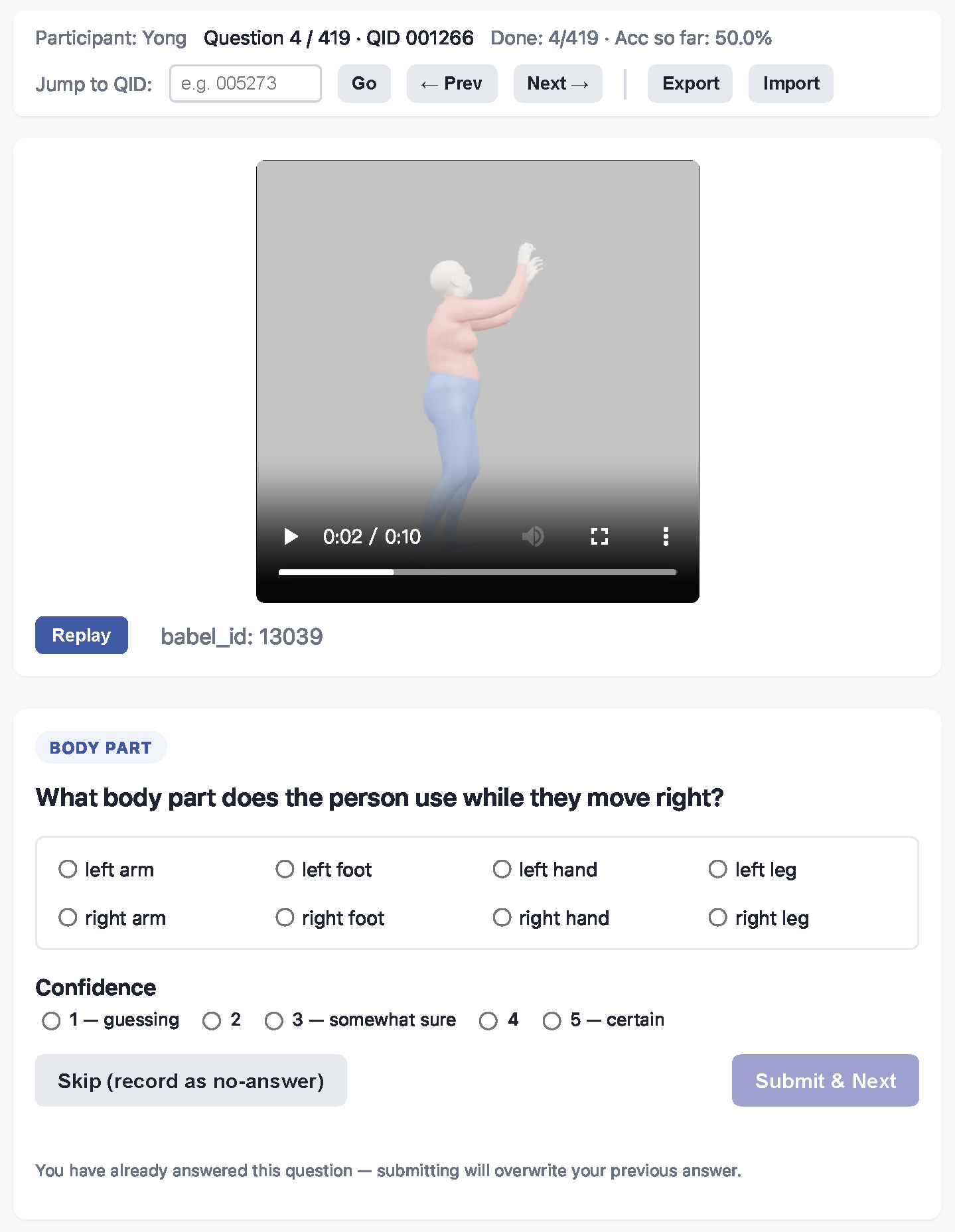}
\caption{In-house annotation interface used in the pilot study. The expert sees the rendered motion clip and the original HumanMotionQA question with its candidate options. This specific example illustrates failure mode F1 (body-part granularity collisions): the gold label is ``right hand'', but the motion exhibits a full-arm reach, causing experts to split between ``right hand'' and ``right arm''.}
\label{fig:human-study-interface}
\end{figure}

Through a rigorous, manual joint-review of the items where the experts disagreed with the gold labels, we categorized the discrepancies into four recurring, systemic failure modes:

\paragraph{F1: Body-part granularity collisions.}
The underlying BABEL vocabulary often conflates hierarchical body parts (e.g., hand vs. arm; foot vs. leg). When the gold answer specifies ``right hand'' but the video clearly depicts a full-arm extension, both ``right hand'' and ``right arm'' are physically and semantically defensible. Experts frequently split on these choices, as illustrated in the interface example (Figure~\ref{fig:human-study-interface}).

\paragraph{F2: Ambiguous spatial frame of reference.}
Directional queries (e.g., ``move right'') systematically fail to specify the frame of reference. It remains undefined whether the direction refers to the viewer’s right, the actor’s egocentric right, or a fixed allocentric world coordinate. HumanMotionQA inherits this ambiguity from BABEL labels, which were crowd-sourced without a unified spatial convention.

\paragraph{F3: Temporal scope mismatch.}
Sequential questions using temporal connective phrases (e.g., ``What did the person do before action X?'') assume a discrete, non-overlapping boundary between actions. However, human motion is inherently continuous, and frame-level annotations in BABEL often overlap or have fuzzy boundaries, making the exact temporal window for the ``before'' or ``after'' action mathematically and perceptually ill-defined.

\paragraph{F4: Composite-action dominance.}
Many motion clips exhibit simultaneous actions (e.g., walking forward while waving a hand). When a question asks ``What action is the person doing?'', there is no single dominant verb. In these cases, the gold label typically reflects the subjective preference of the original crowd annotator rather than an objective truth, resulting in systemic errors when evaluated by domain experts.

% --- Figure 1 移到了附录中 ---

\begin{figure*}[t]
  \centering
  \includegraphics[width=\linewidth]{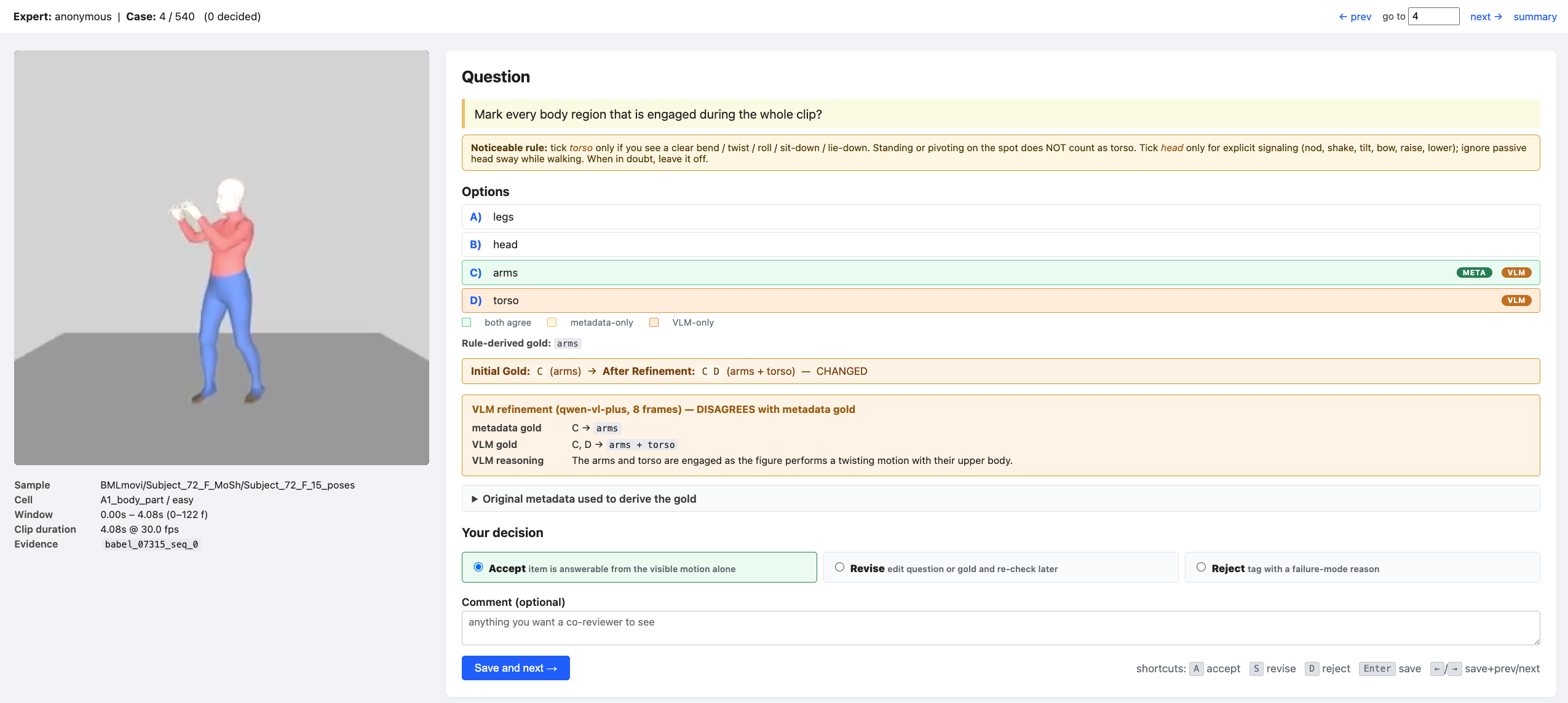}
  \caption{\textbf{Our human annotation platform for NextMotionQA.}
  We build a dedicated web interface to collect human-motion question-answering data.
  For each motion clip, annotators inspect the SOTA-VLM-proposed QA / caption / correction instance and choose one of three actions: \emph{accept}, \emph{revise}, or \emph{reject}.
  This three-way verification is applied to every instance in the benchmark, ensuring annotation quality while keeping human effort tractable.}
  \label{fig:annotation_interface_t1}
\end{figure*}

\begin{figure*}[t]
  \centering
  \includegraphics[width=0.75\linewidth]{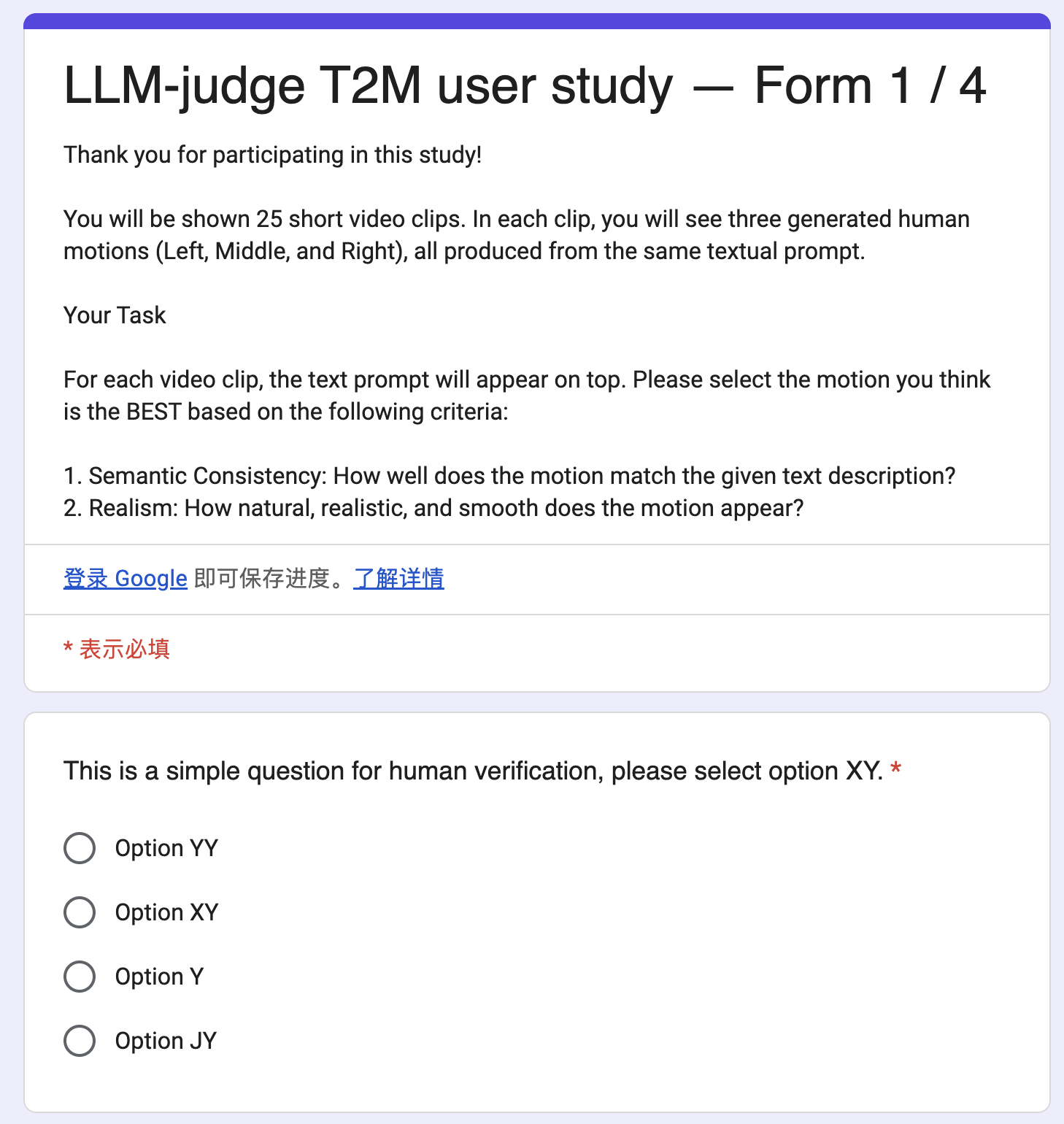}
  \caption{Crowd-sourced user study forms with clear guidelines and evaluation metrics. We use Prolific (\url{https://www.prolific.com/}) as the study platform and recruit qualified participants whose first language is English and who pass our comprehension test.}
  \label{fig:annotation_interface_t1}
\end{figure*}

\section{Evaluated Systems}
\label{ax:evaluated_models}

Table~\ref{tab:model-list} lists the 12 VLMs evaluated in \S\ref{sec:exp-main}, grouped by family and annotated with release version, parameter count where disclosed, and
the access modality used in our experiments. Proprietary
VLMs were queried through their official APIs; open-source VLMs
were run from public weights on a single NVIDIA H100
(80\,GB) node. Motion-specific LLMs were run from the
authors' released checkpoints in their native SMPL input
mode without any fine-tuning on \benchname{}.

\begin{table*}[h]
\centering
\small
\setlength{\tabcolsep}{6pt}
\renewcommand{\arraystretch}{1.1}
\begin{tabular}{llllc}
\toprule
\textbf{Family} & \textbf{Model} & \textbf{Version / Date} & \textbf{Params} & \textbf{Input} \\
\midrule
\multirow{3}{*}{Proprietary VLM}
 & GPT-5.4-mini       & {gpt-5.4-mini-2026-03}     & --                & video \\
 & Qwen3.6-Plus       & {qwen3.6-plus-2026-04}     & --                & video \\
 & Gemini-3.1-Flash   & {Gemini-3.1-flash-2025-09} & --                & video \\
\midrule
\multirow{9}{*}{Open-source VLM}
 & Qwen3.5-0.8B    & {Qwen3.5-0.8B} & 0.8\,B            & video \\
 & Qwen3.5-4B      & {Qwen3.5-4B}   & 4\,B              & video \\
 & Qwen3.5-9B      & {Qwen3.5-9B}   & 9\,B              & video \\
 & Qwen3.5-27B     & {Qwen3.5-27B}  & 27\,B             & video \\
 & InternVL3.5-4B     & {InternVL3.5-4B}           & 4\,B              & video \\
 & InternVL3.5-8B     & {InternVL3.5-8B}           & 8\,B              & video \\
 & InternVL3.5-14B    & {InternVL3.5-14B}          & 14\,B             & video \\
 & 
LLaVA-OneVision-1.5-4B-Instruct     & {LLaVA-OneVision-1.5-4B-Instruct }   & 4\,B              & video \\
 & LLaVA-OneVision-1.5-8B-Instruct      & {LLaVA-OneVision-1.5-8B-Instruct }   & 8\,B              & video \\
\bottomrule
\end{tabular}
\caption{Evaluated VLMs in \benchname{}. All systems are queried with rendered RGB video (at $f$\,FPS and $H{\times}W$ resolution). Proprietary parameter counts are undisclosed.}
\label{tab:model-list}
\end{table*}

\section{VLM-as-A-Judge Evaluation Methods.}
\label{ax:vlm_judges_baseline}

\begin{figure}[t]
\centering
\includegraphics[width=\columnwidth]{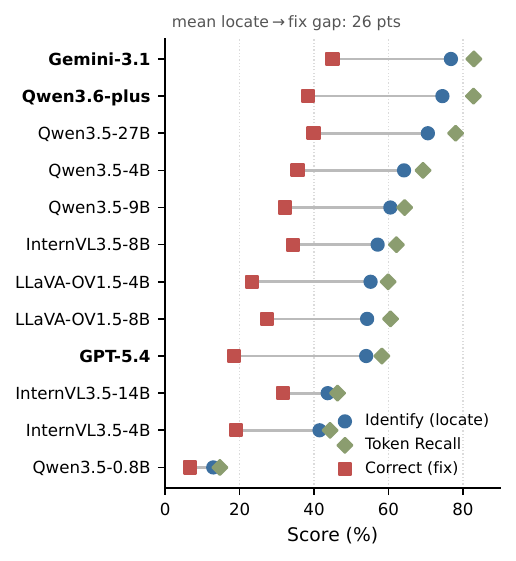}
\caption{\textbf{On T3, models locate motion-caption errors far better
than they fix them.} For each system we show the fraction of gold error
spans \emph{located} (Identify, \textcolor{c_id}{$\bullet$}), the
token-level \emph{recall} of those spans (\textcolor{c_rec}{$\blacklozenge$}),
and the rate at which located errors are semantically \emph{corrected}
(Correct, \textcolor{c_cor}{$\blacksquare$}; judged by an independent
Gemini-3.1-Flash evaluator that accepts synonyms). Systems are sorted by
Identify; closed-source models are in \textbf{bold}. The
locate$\rightarrow$fix gap averages \textbf{26 points} and never closes:
even the best locator, Gemini-3.1-Flash, recovers $76.8\%$ of error spans but
corrects only $44.9\%$ of them.}
\label{fig:t3_funnel}
\end{figure}
  
We evaluate VLM-as-A-Judge on nine representative T2M methods spanning four lines of work. From the diffusion family, we include MDM~\citep{tevet2023mdm}, which establishes continuous motion synthesis, and MotionLCM~\citep{dai2024motionlcm}, a consistency-model variant that improves sampling efficiency. From the discrete-token family, we include MoMask~\citep{guo2024momask}, which casts motion as a language-like sequence. From the continuous-latent family, we include MotionStreamer~\citep{xiao2025motionstreamer}, which introduces continuous causal latents with autoregressive diffusion; MARDM~\citep{meng2025rethinking}, which performs bidirectional masked autoregression in continuous space; and ActionPlan~\citep{nazarenus2026actionplan}, which uses per-frame text-latent anchors with latent-specific diffusion timesteps. From the part-based family, we include FineMoGen~\citep{zhang2023finemogen}, CoMo~\citep{huang2024controllable}, and FrankenMotion~\citep{li2026frankenmotion}, which decompose motion into per-part units guided by independent part-level textual prompts.

\section{Data and Rendering Details}
\label{app:data}

\paragraph{Clip selection.}
We retain AMASS sub-clips of duration
$L\in[0.81, 119.92]$
seconds, after concatenating consecutive BABEL segments
that share an action label. Clips outside this range
either lack the compositional structure required for the
Medium / Hard tiers (too short) or exceed the temporal
window reliably perceived by current video VLMs (too long).
$N_\text{candidate}=1061$ clips passed the duration
filter and entered the difficulty-assignment step.

\paragraph{Difficulty assignment.}
We assign difficulty based on the cardinality of BABEL
frame-level action labels and the modifier content of the
HumanML3D caption. \textbf{Easy:} exactly one BABEL
action label and a HumanML3D caption with no directional,
speed, or fine-body-part modifiers. \textbf{Medium:} two
adjacent BABEL labels with a fast transition
(transition duration $< 0.5$\,s). \textbf{Hard:}
three or more BABEL labels \emph{or} a caption containing
modifiers from the lexicons listed in
Table~\ref{tab:modifier-lex}. Borderline cases were
adjudicated by an expert during Stage 3 verification
(\S\ref{sec:pipeline}).

\begin{table}[h]
\centering
\small
\begin{tabular}{ll}
\toprule
\textbf{Modifier type} & \textbf{Lexicon (examples)} \\
\midrule
Direction & forward, backward, left, right, lateral \\
Speed     & slow, slowly, fast, quickly, briskly \\
Manner    & zigzag, in place, on the spot, in circles \\
Body part & belly, knee, elbow, hand, foot \\
\bottomrule
\end{tabular}
\caption{Modifier lexicons used for Hard-tier assignment.
A clip is upgraded to Hard if its HumanML3D caption
contains at least one entry from any of the listed
categories.}
\label{tab:modifier-lex}
\end{table}

\begin{table*}[t]
\centering
\small
\renewcommand{\arraystretch}{1.25}
\setlength{\tabcolsep}{5pt}
\begin{tabularx}{\textwidth}{@{}l >{\raggedright\arraybackslash}X >{\raggedright\arraybackslash}X >{\raggedright\arraybackslash}X@{}}
\toprule
 & \textbf{A1: Body-part} & \textbf{A2: Direction} & \textbf{A3: Action} \\
\midrule

\textbf{T1}: QA
& \emph{Q: Which body parts does the person use to kick?}\newline
  (A)~\gold{legs}\, (B)~head\, (C)~arms\, (D)~torso
& \emph{Q: How is the person translating in the clip?}\newline
  (A)~forward\, (B)~in place\, (C)~\gold{lateral}\, (D)~backward
& \emph{Q: Mark every body region that ... between 6.327 and 26.196 seconds?}\newline
  (A)~\gold{legs}\, (B)~arms\, (C)~\gold{torso}\, (D)~head \\
\addlinespace[2pt]\midrule

\textbf{T2}: Caption
& A person uses their \gold{legs} to walk forward for a few steps.
& The person moves their legs \gold{forward} to perform a kick.
& The person uses their legs to \gold{walk in place}, then turns and stands. \\
\addlinespace[2pt]\midrule

\textbf{T3}: Correction
& The person performs a kick in place, using their \err{arms}\,\fix{legs} to execute the action.
& The person is doing jumping jacks \err{forward}\,\fix{in place}, moving their legs apart and together.
& The person is \err{walking}\,\fix{crawling} in place, using their arms and legs to move. \\

\bottomrule
\end{tabularx}
\caption{\textbf{Representative NextMotionQA examples} across the three tasks
(rows) and three semantic axes (columns). T1 lists four options with the
\gold{correct} one highlighted; T2 shows the reference caption with the
axis-relevant attribute \gold{highlighted}; T3 shows a corrupted caption with
the localized edit (\err{wrong}\,$\rightarrow$\,\fix{correct}) on the target
axis. Each cell is one canonical instance; the full benchmark stratifies every
(task, axis) cell across Easy/Medium/Hard difficulty.}
\label{tab:examples_ax}
\end{table*}

\paragraph{Rendering parameters.}
Each clip is rendered with a neutral-gender SMPL-H body mesh
  (vertex-coloured by body region) under two camera-mounted directional
  lights and a front-facing static camera that auto-frames the motion
  trajectory. We render with the pyrender offscreen rasteriser (EGL/OpenGL)
  at $f=30$~FPS and pipe frames to FFmpeg. Clip duration follows the
  queried temporal window rather than a fixed length, ranging from
  {0.8} to {119.9}~seconds (mean {10.2}~s). Mean render
  time was {$\approx$2.5}~seconds per clip.
  
\section{Construction Pipeline Prompts}
\label{ax:all_prompt_settings}

We list the three Qwen2.5-VL prompts used by the drafter
in Stage~1 of the pipeline (\S\ref{sec:pipeline}). All
prompts share the same preamble that supplies the
rendered video, the BABEL action labels, and the
HumanML3D caption as auxiliary conditioning.

  Figure~\ref{fig:t3_funnel} decomposes T3 into {localization}
  (Identify and Recall) and {repair} (Correct), exposing a robust gap:
  every one of the twelve systems identifies error spans far more reliably
  than it fixes them, with a mean locate-to-fix drop of $26$ points. The
  pattern holds across the capability spectrum. The strongest locator, Gemini-3.1-Flash, finds $76.8\%$ of gold spans but rewrites only $44.9\%$
  correctly, while Qwen3.6-Plus ($74.5\%$ to $38.4\%$) and Qwen3.5-27B
  ($70.6\%$ to $39.9\%$) show the same collapse. Crucially, the ranking by
  repair differs from the ranking by localization: several closed-source
  systems that top the Identify axis are overtaken on Correct by mid-scale
  open models such as Qwen3.5-27B and InternVL3.5-8B, and GPT-5.4 locates
  competitively ($54.0\%$) yet corrects the least among capable systems
  ($18.4\%$). This dissociation indicates that T3 difficulty stems not from
  {detecting} what is wrong, a recognition skill VLMs already possess,
  but from {producing} a faithful, minimally edited correction, a
  generation skill that does not track model scale or the open versus closed
  boundary. It mirrors the recognize, describe, critique hierarchy of
  \S\ref{sec:construction}: critique requires grounded generation, and that is where
  current systems remain weakest.

\paragraph{Decoding.}
  Open-source models are decoded greedily (\texttt{do\_sample{=}False},
  i.e.\ temperature $0$), with a response budget of $256$ tokens.
  Closed-source API models use each provider's default sampling and a budget
  of $512$ tokens (Gemini-3.1-Flash, Qwen3.6-Plus) or $2048$ tokens
  (GPT-5.4-mini, to accommodate its hidden reasoning trace), with reasoning
  effort set to its minimum where configurable. No length, frequency, or
  presence penalties are applied. Budgets are uniform across the three tasks;
  outputs are far shorter in practice.

\section{Pilot Study and Expert Annotation Interface}
\label{app:interface}

The pilot study (\S\ref{sec:pilot}) and the Stage-3
verification of \benchname{} (\S\ref{sec:pipeline}) share
a single in-house web interface (Figure~\ref{fig:human-study-interface} and Figure~\ref{fig:annotation_interface_t1}).
The interface enforces three properties: (i) the gold
answer is never displayed; (ii) other annotators'
answers are never displayed; and (iii) all motion videos
are auto-looping with a replay button, so an annotator's
decision is never constrained by a single watch.

\paragraph{Annotator instructions.}
Each annotator received a written guideline document
(1~page) covering: the four failure modes
F1--F4, the definition of each semantic axis, the
difficulty rubric, and worked examples of accept / revise
/ reject decisions. Annotators completed
$N_\text{train}=1$ training item with feedback
from the lead author before beginning the production
annotation; training items were not included in the
released benchmark.

\paragraph{Compensation.}
Annotators were salaried researchers compensated as part
of their normal employment; no additional per-item
compensation was paid. The annotation effort totalled
approximately 50~person-minutes across the three
annotators.

\section{Per-Cell Examples}
\label{app:cell-examples}

For completeness we list one representative item per
(task, axis, difficulty) cell in
Table~\ref{tab:examples_ax}. The full benchmark is
released in the supplementary material.

\begin{table}[t]
\centering
\small
\resizebox{0.48\textwidth}{!}{%
\setlength{\tabcolsep}{6pt}
\begin{tabular}{lcccccc}
\toprule
\textbf{Model} & \textbf{A1} & \textbf{A2} & \textbf{A3} & \textbf{Easy} & \textbf{Medium} & \textbf{Hard} \\
\midrule
LLaVA-4B     & 63.33 & 42.44 & 47.78 & 54.55 & 48.52 & 51.02 \\
LLaVA-8B     & 62.18 & 47.09 & 47.73 & 56.10 & 49.58 & 51.71 \\
\midrule
Intern-4B     & 44.26 & 46.80 & 51.20 & 47.67 & 48.01 & 46.62 \\
Intern-8B     & 48.84 & 43.80 & 51.67 & 52.33 & 46.30 & 46.06 \\
Intern-14B    & 45.69 & 29.84 & 38.06 & 41.47 & 38.10 & 34.54 \\
\midrule
Qwen3.5-0.8B       & 40.88 & 23.69 & 37.69 & 38.52 & 32.36 & 32.04 \\
Qwen3.5-4B         & 40.79 & 42.20 & 49.49 & 45.40 & 45.42 & 41.81 \\
Qwen3.5-9B         & 46.48 & 51.07 & 46.44 & 46.08 & 52.45 & 45.23 \\
Qwen3.5-27B        & 46.30 & \textbf{57.07} & 64.03 & 62.79 & 56.25 & 48.61 \\
\midrule
Qwen3.6-plus & \textbf{72.36} & 51.74 & \textbf{68.89} & \textbf{73.35} & \textbf{64.77} & \textbf{55.83} \\
\bottomrule
\end{tabular}}
\caption{Jaccard accuracy (\%) on different attribute dimensions (A1--A3) and difficulty levels (Easy/Medium/Hard). \textbf{Bold} indicates the best result in each column.}
\label{tab:jaccard_t1_ax}
\end{table}
  
% \begin{table}[h]
% \centering
% \small
% \resizebox{0.5\textwidth}{!}{
% \begin{tabular}{lc}
% \toprule
% \textbf{Perturbation}                  & \textbf{System $r$ (vs.\ expert)} \\
% \midrule
% Default (Gemini-3.1-Pro)               & \fillme{0.XX} \\
% \quad + shuffled criterion order       & \fillme{0.XX} \\
% \quad + alternative rubric phrasing    & \fillme{0.XX} \\
% Independent judge  & \fillme{0.XX} \\
% \bottomrule
% \end{tabular}}
% \caption{System-level robustness of the VLM-as-judge
% protocol. Correlations remain within $\pm 0.0\fillme{X}$
% of the default configuration across all perturbations.}
% \label{tab:judge-robust}
% \end{table}

% \paragraph{API cost.}
% Querying the four proprietary VLMs across the full
% benchmark and the VLM-as-judge generation suite cost
% approximately \fillme{\$X{,}XXX} in total at
% \fillme{2026-MM} pricing. Per-request cost ranges from
% \fillme{\$0.0XX} (Gemini-2.0-Flash) to
% \fillme{\$0.XX} (Claude-Sonnet-4.5) per video item.

\section{Ethical Considerations and Licensing}
\label{app:ethics}

\paragraph{Source data.}
AMASS, BABEL, and HumanML3D are publicly released under
research licenses prohibiting commercial use; our release
inherits the most restrictive of the three. No personally
identifying biometric information beyond anonymous SMPL
parameters is redistributed; all rendered videos use a
neutral synthetic body mesh, not a captured subject's
appearance.

\paragraph{Annotation.}
The three expert annotators are co-authors of this
submission and consented to the use of their judgments;
no third-party crowdworkers were employed.

\paragraph{Potential misuse.}
A calibrated VLM-as-judge could in principle be used to
optimize generation systems against the judge rather than
against true motion quality.

\input{prompts_used}

%% file: prompts_used.tex
\begin{figure*}[t]
\centering
\begin{tcblisting}{
    enhanced,
    breakable,
    colframe=promptbox_color_1,
    colback=white,
    coltitle=white,
    colbacktitle=promptbox_color_1,
    rounded corners,
    arc=1.5mm,
    boxrule=0.6mm,
    frame style={solid},
    fonttitle=\bfseries,
    title={Task 1 (QA) -- common preamble.},
    listing only,
    listing options={
        basicstyle=\small\ttfamily,
        breaklines=true,
        breakatwhitespace=false,
        columns=fullflexible,
        keepspaces=true,
        upquote=true,
    }
}
You are an expert annotator for a 3D human-motion benchmark. You will be given the textual annotations describing one motion clip (BABEL action labels and HumanML3D / KIT-ML captions). Your task is to draft a single multiple-choice question item that is answerable purely from observing the visible motion.
The gold (correct) answer for this item is already decided and given to you below. You must NOT change the gold answer. Your only outputs are:
  (1) a one-sentence question stem;
  (2) three plausible, *wrong* distractor options that lie in the same semantic axis as the gold.
CLIP METADATA
-------------
sample_key: {sample_key}
duration:   {duration}s
difficulty: {difficulty}
axis:       {axis}
gold:       {gold}
visible_window_sec: [{scope_start}, {scope_end}]
gold_refers_to_segment: "{dom_text}" [{dom_start}, {dom_end}]s
-- BABEL sequence labels (whole-clip action) --
{babel_seq_block}
-- BABEL segment labels (time-ordered, [start,end] seconds) --
{babel_seg_block}
-- HumanML3D captions --
{humanml3d_block}
-- KIT-ML captions --
{kitml_block}
DIFFICULTY-DEPENDENT PHRASING RULES
-----------------------------------
- difficulty=easy: the visible window IS the whole clip and contains a single action. Phrase the question DIRECTLY without temporal connectives.
- difficulty=medium / hard: the visible window contains multiple actions. The gold refers to the gold_refers_to_segment quoted above. Your question MUST disambiguate which sub-phase you mean by **citing the explicit time range in seconds** (e.g. "between 1.20 and 3.50 seconds"). DO NOT use the phrases "second sub-action", "third sub-action", "first phase", "next phase" -- sub-action segmentation is hard to define visually. Use only explicit seconds + the action name when it is unambiguous in the metadata (e.g. "during the crawl-forward phase between 2.9 and 17.07 seconds").
\end{tcblisting}
\end{figure*}

\begin{figure*}[t]
\centering
\begin{tcblisting}{
    enhanced,
    breakable,
    colframe=promptbox_color_1,
    colback=white,
    coltitle=white,
    colbacktitle=promptbox_color_1,
    rounded corners,
    arc=1.5mm,
    boxrule=0.6mm,
    frame style={solid},
    fonttitle=\bfseries,
    title={Task 1 -- A1 body-part task.},
    listing only,
    listing options={
        basicstyle=\small\ttfamily,
        breaklines=true,
        breakatwhitespace=false,
        columns=fullflexible,
        keepspaces=true,
        upquote=true,
    }
}
AXIS: A1_body_part. The closed answer space is exactly four atomic, non-overlapping body regions: {options_csv}. Some clips legitimately involve MULTIPLE regions (e.g. crawl -> arms + legs; push-up -> arms + torso) -- this question is a "select all that apply" item.

The rule-derived gold body-part SET for this clip is: {gold_set}.
You MUST keep this gold set intact (mark every gold option as correct, and only those). You only choose:
  (1) the question stem, and
  (2) nothing else for the options -- all four atomic regions appear as options A-D; the orchestrator decides which letter each goes into.

REQUIRED question opener: "{opener}". Use it verbatim, then complete with a temporal qualifier matching the difficulty rule above. End with "?". Do NOT use the phrase "primarily involved" -- multiple parts can be correct; phrase the question to ask which parts are involved (select all that apply).
\end{tcblisting}
\end{figure*}

\begin{figure*}[t]
\centering
\begin{tcblisting}{
    enhanced,
    breakable,
    colframe=promptbox_color_1,
    colback=white,
    coltitle=white,
    colbacktitle=promptbox_color_1,
    rounded corners,
    arc=1.5mm,
    boxrule=0.6mm,
    frame style={solid},
    fonttitle=\bfseries,
    title={Task 1 -- A2 direction task.},
    listing only,
    listing options={
        basicstyle=\small\ttfamily,
        breaklines=true,
        breakatwhitespace=false,
        columns=fullflexible,
        keepspaces=true,
        upquote=true,
    }
}
AXIS: A2_direction. Closed answer space (single-correct): {options_csv}.

REQUIRED question opener: "{opener}". Use this opener verbatim at the start of the question, then complete the sentence with a temporal qualifier matching the difficulty rule above. End with "?".

The three distractors MUST be drawn verbatim from {options_csv}, MUST exclude the gold ({gold}), and MUST be distinct from one another. Do not introduce viewer-vs-actor frame distinctions.
\end{tcblisting}
\end{figure*}

\begin{figure*}[t]
\centering
\begin{tcblisting}{
    enhanced,
    breakable,
    colframe=promptbox_color_1,
    colback=white,
    coltitle=white,
    colbacktitle=promptbox_color_1,
    rounded corners,
    arc=1.5mm,
    boxrule=0.6mm,
    frame style={solid},
    fonttitle=\bfseries,
    title={Task 1 -- A3 action task.},
    listing only,
    listing options={
        basicstyle=\small\ttfamily,
        breaklines=true,
        breakatwhitespace=false,
        columns=fullflexible,
        keepspaces=true,
        upquote=true,
    }
}
AXIS: A3_action. The gold action label is "{gold}".

REQUIRED question opener: "{opener}". Use this opener verbatim at the start of the question, then complete with a temporal qualifier matching the difficulty rule above. End with "?".

Propose THREE distractors. CONSTRAINTS:
  - Each distractor MUST be drawn from this curated pool of common human-motion verbs (or be a very close verb-phrase variant of an entry):
{distractor_pool}
  - Distractors MUST come from a DIFFERENT VERB FAMILY than the gold. For example, if the gold is "jog" (locomotion-normal), do NOT pick "run" or "march" (same family -- too close); pick something like "kick", "wave", or "crawl" (different families).
  - Distractors MUST NOT be near-synonyms of each other (no "run"/"jog", no "look"/"glance"/"stare"/"gaze", no "jump"/"hop"/"leap" together).
  - Distractors MUST NOT be rare or unusual verbs (no "lurch", "saunter", "trundle", "career"). Stick to verbs in the pool above.
  - 1-4 words each.
\end{tcblisting}
\end{figure*}

\begin{figure*}[t]
\centering
\begin{tcblisting}{
    enhanced,
    breakable,
    colframe=promptbox_color_1,
    colback=white,
    coltitle=white,
    colbacktitle=promptbox_color_1,
    rounded corners,
    arc=1.5mm,
    boxrule=0.6mm,
    frame style={solid},
    fonttitle=\bfseries,
    title={Task 1 -- output format.},
    listing only,
    listing options={
        basicstyle=\small\ttfamily,
        breaklines=true,
        breakatwhitespace=false,
        columns=fullflexible,
        keepspaces=true,
        upquote=true,
    }
}
OUTPUT FORMAT
-------------
Return ONLY a single JSON object (no prose, no markdown fences) with exactly these fields:

{
  "question":    "...",
  "distractors": ["...", "...", "..."]    # ignored for A1
}

Constraints:
  - "question" is a single sentence; <= 30 words; ends with a "?".
  - "distractors" has exactly 3 strings, each strictly different from the gold ({gold}). For A1 the orchestrator fills the four options from the closed lexicon -- leave "distractors" as an empty list [] or any 3 placeholder strings; they will be discarded.
\end{tcblisting}
\end{figure*}

\begin{figure*}[t]
\centering
\begin{tcblisting}{
    enhanced,
    breakable,
    colframe=promptbox_color_2,
    colback=white,
    coltitle=white,
    colbacktitle=promptbox_color_2,
    rounded corners,
    arc=1.5mm,
    boxrule=0.6mm,
    frame style={solid},
    fonttitle=\bfseries,
    title={Task 2 (captioning) drafting prompt.},
    listing only,
    listing options={
        basicstyle=\small\ttfamily,
        breaklines=true,
        breakatwhitespace=false,
        columns=fullflexible,
        keepspaces=true,
        upquote=true,
    }
}
You are an expert annotator for a 3D human-motion benchmark. You write rich reference captions that will be used to grade an open-vocabulary video-captioning task. Captions must be answerable purely from a short rendered video of one person moving.

CLIP METADATA
-------------
sample_key:  {sample_key}
duration:    {duration}s
difficulty:  {difficulty}
visible_window_sec: [{scope_start}, {scope_end}]
gold_refers_to_segment: "{dom_text}" [{dom_start}, {dom_end}]s

ELIGIBLE AXES + GOLDS (the caption MUST surface every one of these)
{axes_block}

-- BABEL action labels --
{babel_block}

-- HumanML3D captions --
{humanml3d_block}

TASK
----
Write a reference caption (1-2 sentences, 30-40 words) that describes the visible motion and **explicitly mentions every gold attribute listed above**. The caption must:

  - be answerable from the visible motion alone (no objects the renderer cannot show -- no balls, phones, instruments, etc);
  - be in natural English (not a bullet list, not a template);
  - contain a contiguous substring that exactly matches EACH gold attribute (verbatim, lower-case, no quotes around it). For multi-element A1 golds like "arms + legs", you can either put both in one phrase (e.g. "using their arms and legs") OR mention each part separately in the same sentence -- but the substring "arms" and the substring "legs" must each appear somewhere in the caption.

Then identify the CHARACTER OFFSETS [start, end) of each gold attribute substring inside your caption. Offsets are 0-indexed half-open; caption[start:end] must equal the substring exactly.

OUTPUT
------
Return ONLY a single JSON object with exactly these fields:

  {
    "reference_caption": "<one or two sentences>",
    "attribute_spans": [
      {"axis": "<one of {axes_csv}>",
        "start": <int>, "end": <int>,
        "text": "<exact substring of the caption>"},
      ...
    ]
  }

Constraints:
  - attribute_spans contains EXACTLY one entry per eligible axis listed above (so for {n_axes} eligible axes you return {n_axes} entries).
  - For multi-element A1 golds (e.g. "arms + legs"), still ONE entry: pick a phrase that covers both elements (e.g. "arms and legs") OR mark only the primary element if a contiguous phrase isn't possible.
\end{tcblisting}
\end{figure*}

\begin{figure*}[t]
\centering
\begin{tcblisting}{
    enhanced,
    breakable,
    colframe=promptbox_color_3,
    colback=white,
    coltitle=white,
    colbacktitle=promptbox_color_3,
    rounded corners,
    arc=1.5mm,
    boxrule=0.6mm,
    frame style={solid},
    fonttitle=\bfseries,
    title={Task 3 (error-correction) corruption prompt.},
    listing only,
    listing options={
        basicstyle=\small\ttfamily,
        breaklines=true,
        breakatwhitespace=false,
        columns=fullflexible,
        keepspaces=true,
        upquote=true,
    }
}
You are an expert annotator producing items for a 3D human-motion error-correction benchmark. You are given a KNOWN-CORRECT reference caption R that already mentions the gold attributes for one or more semantic axes. Your job is to produce a CORRUPTED caption C with **ONE TO THREE disjoint, localized lexical errors**, and a REWRITE that restores all gold attributes.

CLIP METADATA
-------------
sample_key:    {sample_key}
eligible_axes: {axes_csv}
gold_by_axis:  {gold_table}

REFERENCE CAPTION R
-------------------
"{reference}"

Reference attribute spans (the parts of R that name each gold; you may corrupt one or more of these):
{attr_spans_block}

CORRUPTION RULES
----------------
{per_axis_rules}

  - Produce ONE to THREE disjoint error spans in C. Each span:
      * is contiguous in C (no nested spans);
      * covers <= 8 words;
      * is a real, on-axis error -- its replaced phrase does NOT match the gold (and is not a near-synonym);
      * belongs to ONE of the eligible axes; different spans MAY belong to DIFFERENT eligible axes (e.g. one A2 error AND one A1 error in the same item).
  - All non-error characters in C -- including punctuation, casing, whitespace -- MUST be identical to R. Length differences between R and C arise only inside the error spans.
  - Each error_span's char offsets [start, end) are computed against the *corrupted* caption C and must satisfy C[start:end] == text.
  - The "gold_text" field of each span is the corresponding correct phrase from R (i.e. what the rewrite restores).
  - The REWRITE is a single sentence (or two) that restores every gold attribute. Usually `rewrite == R` is fine.

DIFFICULTY HINT
---------------
Aim for variety: roughly a third of items with 1 error, a third with 2, a third with 3. Multi-axis combinations (e.g. an A2 error + an A1 error) are preferred when the clip is eligible for >=2 axes.

OUTPUT
------
Return ONLY a single JSON object with exactly these fields:

  {
    "corrupted_caption": "<one or two sentences>",
    "error_spans": [
      {"axis": "<one of {axes_csv}>",
        "start": <int>, "end": <int>,
        "text": "<wrong substring of C>",
        "gold_text": "<the correct phrase that should be there>"},
      ...   # 1 to 3 entries
    ],
    "rewrite": "<one or two sentences restoring every gold>"
  }
\end{tcblisting}
\end{figure*}

\begin{figure*}[t]
\centering
\begin{tcblisting}{
    enhanced,
    breakable,
    colframe=promptbox_color_3,
    colback=white,
    coltitle=white,
    colbacktitle=promptbox_color_3,
    rounded corners,
    arc=1.5mm,
    boxrule=0.6mm,
    frame style={solid},
    fonttitle=\bfseries,
    title={Task 3 -- per-axis corruption rules.},
    listing only,
    listing options={
        basicstyle=\small\ttfamily,
        breaklines=true,
        breakatwhitespace=false,
        columns=fullflexible,
        keepspaces=true,
        upquote=true,
    }
}
[A1_body_part] swap the body-part phrase for a DIFFERENT atomic body region from {head, arms, legs, torso} that does NOT match the gold
[A2_direction] swap the direction phrase for a DIFFERENT direction from {forward, backward, lateral, in place} that does NOT match the gold
[A3_action] swap the action verb-phrase for a DIFFERENT common motion verb (walk / run / jump / hop / kick / throw / wave / clap / point / sit / stand / lie / crawl / climb / push / pull / lift / bow / stretch / turn around / dance / march) that does NOT match the gold
\end{tcblisting}
\end{figure*}

\begin{figure*}[t]
\centering
\begin{tcblisting}{
    enhanced,
    breakable,
    colframe=promptbox_color_1,
    colback=white,
    coltitle=white,
    colbacktitle=promptbox_color_1,
    rounded corners,
    arc=1.5mm,
    boxrule=0.6mm,
    frame style={solid},
    fonttitle=\bfseries,
    title={Task 1 VLM refinement -- header.},
    listing only,
    listing options={
        basicstyle=\small\ttfamily,
        breaklines=true,
        breakatwhitespace=false,
        columns=fullflexible,
        keepspaces=true,
        upquote=true,
    }
}
You are an expert annotator for a 3D human-motion benchmark. You will be shown a short video of one person moving (rendered SMPL mesh, red shirt / blue pants, the head turns GREEN during the gold sub-window when one is set). You will also be given a multiple-choice question with four options A, B, C, D.

Your job is to (1) pick the option(s) that correctly answer the question based ONLY on what you see in the motion, and (2) propose a clearer question phrasing IF the current question is ambiguous or uses vague language. Use the person's own body-axes (forward = direction they face; their right is the opposite side of the screen when they face the camera).

QUESTION-PHRASING REVIEW
------------------------
Read the question carefully. If any of the following holds, return a "revised_question" with a clearer rewrite; otherwise return an empty string for that field.

  - vague verbs like "active", "engaged", "involved" without a specific     action anchor (prefer phrasings like "which parts execute the action"     or "major functioning parts for the action shown");
  - placeholder text leaked from the template ("t1", "X-ing", "the X phase");
  - undefined references ("the segment", "this part") with no time anchor;
  - grammar / typo issues that make the question awkward.

When you DO revise: keep the question's intent and temporal scope, only rephrase. Example: "Which atomic body regions are active during the whole clip?" -> "Which body parts are the major functioning parts for the action shown in the whole clip?".

When you do NOT revise, return "revised_question": "".
\end{tcblisting}
\end{figure*}

\begin{figure*}[t]
\centering
\begin{tcblisting}{
    enhanced,
    breakable,
    colframe=promptbox_color_1,
    colback=white,
    coltitle=white,
    colbacktitle=promptbox_color_1,
    rounded corners,
    arc=1.5mm,
    boxrule=0.6mm,
    frame style={solid},
    fonttitle=\bfseries,
    title={VLM refinement -- A1 suffix.},
    listing only,
    listing options={
        basicstyle=\small\ttfamily,
        breaklines=true,
        breakatwhitespace=false,
        columns=fullflexible,
        keepspaces=true,
        upquote=true,
    }
}
AXIS: A1_body_part. This is a "select all that apply" item: the closed answer space is four atomic, non-overlapping body regions (head / arms / legs / torso), and multiple regions may legitimately be involved.

NOTICEABLE-AND-PRIMARY rule (very important -- applies to torso and head):

  - **Torso**: only mark torso if the action's defining motion is a clearly
    visible torso bend, twist, rotation, roll, or sit-down / lie-down. Do NOT
    mark torso just because it is upright, held, or making small balancing
    adjustments while the legs walk / run / spin / turn. A standing,
    balancing, or pivoting person is legs-only.
  - **Head**: only mark head if there is an explicit signaling head motion
    (nod, shake, tilt, bow, raise, lower). Do NOT mark head for passive
    glances, looking around while walking, or incidental head sway. If you
    cannot describe the head motion in one verb, it is not noticeable.

When in doubt, leave the part OFF. Over-marking torso and head was the most
common refine error in the pilot study.

Return ONLY a single JSON object with exactly these fields:

  {
    "answer":           ["A", ...],     // letters of EVERY option that is a noticeable AND primary driver
    "reasoning":        "one short sentence",
    "revised_question": ""               // empty string OR a clearer rewrite
  }
\end{tcblisting}
\end{figure*}

\begin{figure*}[t]
\centering
\begin{tcblisting}{
    enhanced,
    breakable,
    colframe=promptbox_color_1,
    colback=white,
    coltitle=white,
    colbacktitle=promptbox_color_1,
    rounded corners,
    arc=1.5mm,
    boxrule=0.6mm,
    frame style={solid},
    fonttitle=\bfseries,
    title={VLM refinement -- A2 suffix.},
    listing only,
    listing options={
        basicstyle=\small\ttfamily,
        breaklines=true,
        breakatwhitespace=false,
        columns=fullflexible,
        keepspaces=true,
        upquote=true,
    }
}
AXIS: A2_direction. Pick EXACTLY ONE option.

Return ONLY a single JSON object with exactly these fields:

  {
    "answer":           ["A"],          // exactly one letter
    "reasoning":        "one short sentence",
    "revised_question": ""               // empty string OR a clearer rewrite
  }
\end{tcblisting}
\end{figure*}

\begin{figure*}[t]
\centering
\begin{tcblisting}{
    enhanced,
    breakable,
    colframe=promptbox_color_1,
    colback=white,
    coltitle=white,
    colbacktitle=promptbox_color_1,
    rounded corners,
    arc=1.5mm,
    boxrule=0.6mm,
    frame style={solid},
    fonttitle=\bfseries,
    title={VLM refinement -- A3 suffix.},
    listing only,
    listing options={
        basicstyle=\small\ttfamily,
        breaklines=true,
        breakatwhitespace=false,
        columns=fullflexible,
        keepspaces=true,
        upquote=true,
    }
}
AXIS: A3_action. Pick EXACTLY ONE option that best matches what the person is doing in the indicated time range.

Return ONLY a single JSON object with exactly these fields:

  {
    "answer":           ["A"],          // exactly one letter
    "reasoning":        "one short sentence",
    "revised_question": ""               // empty string OR a clearer rewrite
  }
\end{tcblisting}
\end{figure*}